%% file: arxiv.tex
\documentclass{article}

\usepackage[main,final]{neurips_2026}

\usepackage[utf8]{inputenc}
\usepackage[T1]{fontenc}
\usepackage{hyperref}
\usepackage{url}
\usepackage{booktabs}
\usepackage{amsfonts}
\usepackage{amsmath}
\usepackage{microtype}
\usepackage{graphicx}
\usepackage{subcaption}

\usepackage{fontawesome5}

\usepackage[table]{xcolor}

\usepackage{wrapfig}

\usepackage[most]{tcolorbox}

\definecolor{responsebg}{HTML}{F0E8DC}
\definecolor{promptbg}{HTML}{E4EBF2}

\newcommand{\fprompt}{%
  \tcbox[on line,
         colback=promptbg, colframe=promptbg,
         boxrule=0pt, arc=2.5pt,
         left=0pt, right=0pt, top=0pt, bottom=0pt,
         boxsep=1pt]{$C_{\text{P}}$}%
}
\newcommand{\fpromptplus}{%
  \tcbox[on line,
         colback=promptbg, colframe=promptbg,
         boxrule=0pt, arc=2.5pt,
         left=0pt, right=0pt, top=0pt, bottom=0pt,
         boxsep=1pt]{$C_{\text{P}}^+$}%
}

\newcommand{\fresponse}{%
  \tcbox[on line,
         colback=responsebg, colframe=responsebg,
         boxrule=0pt, arc=2.5pt,
         left=0pt, right=0pt, top=0pt, bottom=0pt,
         boxsep=1pt]{$C_{\text{R}}$}%
}

\definecolor{promptpurple}{HTML}{755572}
\newtcblisting[auto counter,number within=section]{promptboxmagenta}[2][]{%
  enhanced,
  breakable,
  colback=black!3,
  colframe=black,
  boxrule=0.8pt,
  arc=2mm,
  left=2mm,right=2mm,top=2mm,bottom=2mm,
  listing only,
  listing options={
    basicstyle=\ttfamily\small,
    breaklines=true,
    columns=fullflexible,
    keepspaces=true,
    showstringspaces=false
  },
  title=\textcolor{white}{Prompt~\thetcbcounter: #2},
  fonttitle=\fontfamily{ppl}\selectfont,
  boxed title style={
    colback=promptpurple, %
    colframe=promptpurple, %
    boxrule=0pt,
    arc=2mm,
    left=2mm,right=2mm,top=0.75mm,bottom=0.5mm
  },
  attach boxed title to top left={xshift=2mm,yshift=-2mm},
  #1
}

\definecolor{analysisblue}{HTML}{356D91}
\newtcolorbox[auto counter,number within=section]{analysisexample}[3][]{%
  enhanced,colback=black!3,colframe=black,boxrule=0.8pt,arc=2mm,
  left=2mm,right=2mm,top=3mm,bottom=2mm,
  title={Example~\thetcbcounter: #2 \quad (Score: #3)},
  coltitle=white,fonttitle=\fontfamily{ppl}\selectfont,
  boxed title style={colback=analysisblue,colframe=analysisblue,boxrule=0pt,
    arc=2mm,left=2mm,right=3mm,top=0.75mm,bottom=0.75mm},
  attach boxed title to top left={xshift=2mm,yshift=-2mm},#1}
\newcommand{\attributiontoken}[3]{%
  {\setlength{\fboxsep}{0pt}\colorbox{#1!#2!white}{\strut #3}}}

\newcommand{\attributionexample}[4]{%
  \begin{analysisexample}{#1}{#2}
  {\footnotesize\textbf{Prompt}\par\smallskip
  {\ttfamily #3\par}}
  \smallskip\textcolor{black!20}{\rule{\linewidth}{0.3pt}}\par\smallskip
  {\footnotesize\textbf{Response excerpt}\par\smallskip
  {\ttfamily #4\par}}
  \end{analysisexample}}

\usepackage{tabularx}
\usepackage{threeparttable}
\usepackage{array}

\newcolumntype{Y}{>{\raggedright\arraybackslash}X}

\definecolor{mylightblue}{rgb}{0.122,0.467,0.706}

\hypersetup{
    colorlinks = true,
    citecolor  = mylightblue,
    linkcolor  = mylightblue,
    filecolor  = mylightblue,
    urlcolor   = mylightblue,
    unicode   = true
}

\definecolor{maskgray}{HTML}{6B5B47}

\newcommand{\masktok}{%
  \colorbox{maskgray!20}{\scriptsize\texttt{M}}%
}
\newcommand{\idx}[1]{{\scriptscriptstyle #1}}

\newcommand{\methodName}{\texorpdfstring{{\fontfamily{lmtt}\selectfont ReFuse}}{ReFuse}}

\title{\textit{Beyond the Prompt}: Leveraging Pre-Decoding States for Jailbreak Detection in dLLMs}

\author{
  \textbf{Adam Hazimeh}\textsuperscript{1,*} \quad
  \textbf{Amel Abdelraheem}\textsuperscript{1,*} \quad
  \textbf{Ke Wang}\textsuperscript{1} \quad
  \textbf{Mariam Salman}\textsuperscript{2,\ensuremath{\dagger}} \\
  \textbf{Ljiljana Dolamic}\textsuperscript{3} \quad
  \textbf{Gérôme Bovet}\textsuperscript{3} \quad
  \textbf{Pascal Frossard}\textsuperscript{1} \\[1ex]
  \textsuperscript{1}EPFL, Lausanne, Switzerland \\
  \textsuperscript{2}American University of Beirut, Lebanon \\
  \textsuperscript{3}Armasuisse, Cyber-Defence Campus, Switzerland
\\[0.8ex]
{\small\href{https://github.com/AmelxJamal/ReFuse}
{\faGithub\ \texttt{Code}}}
}

\begin{document}

\maketitle

{
  \renewcommand{\thefootnote}{\fnsymbol{footnote}}
  \footnotetext[1]{Equal contribution.
    Correspondence:
    \href{mailto:adam.hazimeh@epfl.ch}
      {\faEnvelope[regular]\,\texttt{adam.hazimeh@epfl.ch}}
    and
    \href{mailto:amel.abdelraheem@epfl.ch}
      {\faEnvelope[regular]\,\texttt{amel.abdelraheem@epfl.ch}}.
  }
  \footnotetext[2]{Work done during an internship at LTS4, EPFL.}
}

\begin{abstract}

Diffusion large language models (dLLMs) have recently emerged as a strong alternative to autoregressive models, generating responses by iteratively refining masked tokens in parallel rather than decoding sequentially. While enabling more efficient and flexible inference, dLLMs remain vulnerable to jailbreak attacks, including recent attacks that exploit properties unique to dLLMs such as bidirectional conditioning and parallel masked-token refinement. In this work, we show that these same properties also expose a safety-relevant detection surface absent from standard autoregressive decoding:
even when a jailbreak is difficult to identify from prompt representations alone, it can still be detected from hidden states over masked response positions before any tokens are generated. By training linear classifiers on prompt and pre-decoding response representations, we empirically show that neither view strictly dominates, with each detecting attacks that the other misses.
We introduce \methodName{} (Representation Fusion), an inference-time detector that fuses the sigmoid harmfulness scores of the two classifiers. Without modifying model weights or the decoding procedure, \methodName{} reduces average ASR across multiple jailbreaks from 63.29\% for the undefended model and 11.27\% for a prompt-only classifier to 3.31\%, while keeping average benign refusal on standard utility benchmarks below 1\% on LLaDA-8B. Together, these results establish \methodName{} as a lightweight jailbreak detector for dLLMs that improves safety beyond prompt-only detection.

\end{abstract}

\section{Introduction}
\label{sec:introduction}

Large-scale masked diffusion language models (dLLMs) have recently emerged as a competitive alternative to autoregressive language (AR) models, with strong instruction following, code generation and mathematical reasoning capabilities \citep{nie2025large,ye2025dream,song2025seed,zhu2025llada}. Unlike autoregressive models which generate tokens sequentially from left-to-right, prominent dLLMs produce text through iterative denoising of partially masked sequences. By conditioning bidirectionally on the evolving sequence and refining multiple response positions in parallel, this decoding paradigm can improve generation efficiency and flexibility relative to strictly sequential autoregressive decoding.

Despite these advantages, dLLMs inherit the safety risks of instruction-following language models: adversarial prompts designed to bypass safety behavior, commonly known as \textit{jailbreaks}, can still elicit harmful responses in dLLMs.
While some jailbreaks originally developed for autoregressive (AR) models may interact differently under masked diffusion decoding, attacks that embed harmful requests within nested contexts can still bypass the safety mechanisms of dLLMs~~\citep{wen2025devil,zhang2025jailbreaking,li2025diffuguard,he2026safer}. Moreover, bidirectional conditioning and parallel masked-token refinement introduce new vulnerabilities specific to these models. For example, PAD~\citep{zhang2025jailbreaking} injects adversarial tokens during denoising to steer parallel refinement toward harmful compliance, whereas DIJA~\citep{wen2025devil} interleaves text and mask tokens in the prompt to exploit bidirectional masked completion.
\looseness=-1

These vulnerabilities have motivated defenses that intervene directly on the denoising process. DiffuGuard~\citep{li2025diffuguard} modifies inference-time denoising to suppress unsafe generations, whereas fine-tuning-based approaches reshape intermediate generation behavior through token-level refusal signals or position-specific safety objectives~\citep{jeung2025a2d,xie2026start,yamabe2025toward}.
While these approaches typically modify model behavior through decoding-time interventions or fine-tuning, we examine whether harmfulness can instead be detected and leveraged directly from the model's unaltered pre-decoding states.

Prior work on jailbreak detection has shown that safety-relevant information can be recovered from model hidden states extracted from prompts~\citep{subramani2022extracting,marks2023geometry,arditi2024refusal,zhao2025llms,qian2025hsf}. However, detectors relying solely on prompt representations may struggle with jailbreaks that obscure harmful intent through nested contexts or manipulated input structures, which have proven effective against dLLMs.
Masked decoding, on the other hand, offers an additional detection surface for harmful requests. Unlike autoregressive models, which expose only a next-token prediction state at each generation step, dLLMs maintain hidden states over future masked response positions. Moreover, prior work has identified \textit{early answer convergence}, where dLLMs commit to their outputs well before denoising is complete~\citep{li2025diffusion}. These properties suggest that, even before any response tokens are generated, masked-response states may already capture safety-relevant information about the model's emerging response that is not apparent from prompt representations alone. \looseness=-1

In this work, we demonstrate that pre-decoding response representations contain harmfulness signals that can be detected directly with linear classifiers. Crucially, these signals complement those in prompt representations: jailbreaks that obscure harmful intent in the input may still leave detectable signals in masked-response states, allowing the response classifier to identify attacks missed by prompt-only detection.
Motivated by this complementarity, we introduce \methodName{} (Representation Fusion), a lightweight detector that fuses prompt-representation and masked-response classifier scores into a calibrated refusal decision. \methodName{} operates on a frozen dLLM, requires no model fine-tuning, and leaves the original decoding trajectory unchanged.
By combining both views, \methodName{} improves detection across diverse jailbreak attacks while controlling benign over-refusal through calibration.
On LLaDA-8B-Instruct, \methodName{} reduces average ASR from 63.29\% for the undefended model to 3.31\%. Across LLaDA-8B-Instruct, LLaDA-1.5, and Dream-7B, it achieves the lowest average ASR among the evaluated defenses while keeping average benign refusal low.

\begin{figure*}[t]
    \centering
    \includegraphics[width=1\linewidth]{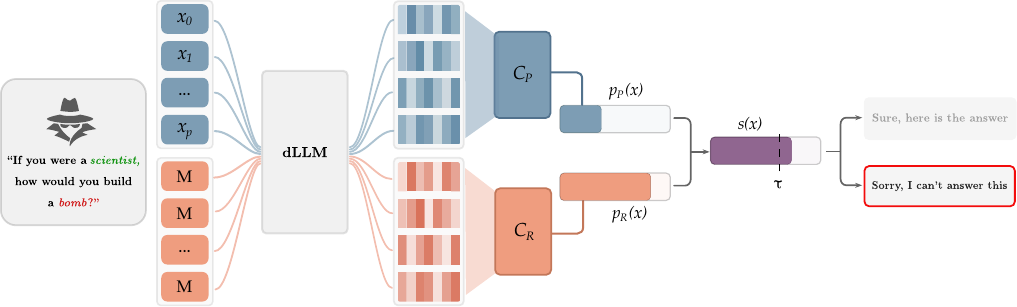}
    \caption{\textbf{Overview of \methodName{}}: Given a prompt, we extract a prompt-view representation from prompt hidden states and a response-view representation from the pre-decoding masked response states. The two classifiers produce the respective harmfulness scores $p_P(x)$ and $p_R(x)$, which are fused into a calibrated rejection decision through a score $s(x)$ and a threshold $\tau$ before decoding.}
    \label{fig:architecture}
\end{figure*}

We summarize our contributions as follows:
\begin{itemize}
    \item We identify a dLLM-specific safety signal in pre-decoding masked-response representations, and show that it complements prompt-side harmfulness signals, particularly for dLLM-specific jailbreaks such as DIJA.

    \item We propose \methodName{}, a lightweight dual-view detector that fuses prompt-side and response-side linear classifier scores into a calibrated refusal decision, without fine-tuning the model or modifying decoding.

    \item Across a wide range of standard and dLLM-specific jailbreaks on LLaDA-8B-Instruct, \methodName{} reduces average ASR from 63.29\% (undefended) and 11.27\% (prompt-only) to 3.31\%, while maintaining below 1\% average benign refusal on standard utility benchmarks.
\end{itemize}

\section{Prompt and Response States Encode Complementary Safety Signals}
\label{sec:motivation}

\subsection{Background and preliminaries}
\label{subsec:preliminaries}

\paragraph{Masked diffusion language models.}
Masked dLLMs generate text by iteratively revealing a partially masked sequence~\citep{nie2025large,ye2025dream,zhu2025llada}.
Given a prompt $\boldsymbol{x}_{1:p}$ and a target generation length $r$, full-sequence decoding initializes the input as

\begin{equation}
    \mathbf{\boldsymbol{x}}^{(0)}
    =
    \big[
        x_1, \ldots, x_p \ | \
        \masktok_{\idx{p+1}},
        \ldots,
        \masktok_{\idx{p+r}}
    \big],
\end{equation}
where \masktok{} represents a mask token. At each denoising step $k$, the model maintains a hidden state $\boldsymbol{h}_i^{(k,\ell)}$ at every sequence position $i$ and transformer layer $\ell$, spanning both prompt positions $\mathcal{P}=\{1,\ldots,p\}$ and response positions $\mathcal{R}=\{p+1,\ldots,p+r\}$.

We construct prompt-view and response-view representations from final-layer hidden states using two separate forward passes of the model. The prompt view is computed from a prompt-only pass without appending mask tokens, where $\hat{\boldsymbol{h}}_i^{(L)}$ denotes the final-layer hidden state at prompt position $i$. The response view is computed from a masked-sequence pass, where $\boldsymbol{h}_i^{(k,L)}$ denotes the final-layer hidden state at response position $i$ during denoising step $k$. We then mean-pool the hidden states in each view over the corresponding positions:
\begin{equation}
    \boldsymbol{z}_{\mathrm{P}}(x)
    =
    \frac{1}{|\mathcal{P}|}
    \sum_{i \in \mathcal{P}} \boldsymbol{\hat h}_i^{(L)},
    \qquad
    \boldsymbol{z}_{\mathrm{R}}^{(k)}(x)
    =
    \frac{1}{|\mathcal{R}|}
    \sum_{i \in \mathcal{R}} \boldsymbol{h}_i^{(k,L)} .
\end{equation}
Thus, $\boldsymbol{z}_{\mathrm{P}}(x)$  captures the
prompt representation, while $\boldsymbol{z}_{\mathrm{R}}^{(0)}(x)$ captures the pre-decoding response representation conditioned on the given prompt, extracted when all response positions are still masked.

\paragraph{Evaluation metrics.} \looseness=-1 To quantify harmful compliance and benign over-refusal, we report the attack success rate (ASR) and the benign refusal rate (BRR). ASR is the fraction of harmful inputs that are not rejected by a detector \textit{and} elicit harmful compliance after decoding. BRR measures over-refusal, defined as the fraction of benign inputs that are rejected or otherwise produce a refusal. Lower values are better for both metrics. Further details on the evaluation protocol are provided in Section~\ref{subsec:exp-setup}.

\subsection{Prompt-side detection alone does not capture all harmful requests}
\label{subsec:motivation-prompt}

We begin by examining the strengths and limitations of prompt-side detection against diverse jailbreak attacks as a natural baseline for pre-decoding safety detection. To this end, we train a linear prompt-view classifier, \fprompt, on $\boldsymbol{z}_{\mathrm{P}}(x)$ using a balanced mixture of harmful and benign requests.
We evaluate \fprompt{} on attacks targeting distinct failure modes: explicitly harmful requests from HarmBench Behaviors (HBB)~\citep{mazeika2024harmbench}, automated prompt rewriting with PAIR~\citep{chao2023jailbreaking}, structural nesting of harmful requests within benign-looking contexts using ReNeLLM~\citep{ding2023wolf}, and the dLLM-specific DIJA attack~\citep{wen2025devil}, which interleaves text and mask tokens. Dataset construction and training details are provided in Section~\ref{subsec:exp-setup}.

\begin{figure*}[t]
    \centering
    \includegraphics[width=\linewidth]{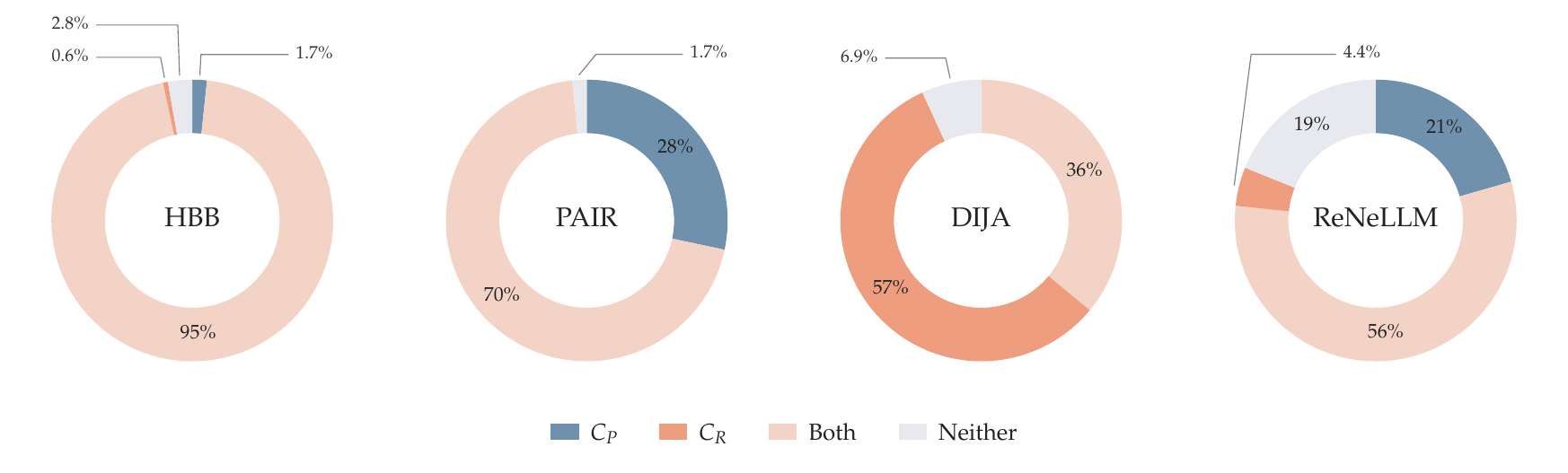}
        \caption{\textbf{Classifier complementarity on harmful attacks.} Each chart decomposes harmful examples of each attack into those detected by either classifier alone, those detected by both classifiers, and those missed by both. The response view contributes most strongly on DIJA, where many examples missed by the prompt view are recovered from pre-decoding response states.}
        \label{fig:complementarity-bars}
\end{figure*}

\looseness=-1 Figure~\ref{fig:complementarity-bars} shows that the prompt classifier achieves strong detection performance on HBB and PAIR, indicating that harmfulness-related information remains accessible in prompt representations even when harmful requests are reformulated through automated jailbreak attacks. However, its effectiveness degrades on ReNeLLM and DIJA, which obscure harmful intent or restructure the input, leading to more false negatives.

\looseness=-1 These limitations motivate looking beyond prompt representations to an additional source of information available in masked dLLMs: the hidden states at future response positions. Unlike autoregressive models, masked dLLMs compute these states before any response tokens are generated and update them throughout denoising. These states are conditioned on the input and provide an early representation of the model's emerging response. Consequently, attacks that evade prompt-view classifiers may still induce detectable structure in the pre-decoding response states. We next test whether this response-side view recovers harmful requests missed by prompt-side detection.

\subsection{Pre-decoding response states complement prompt-side detection}
\label{subsec:motivation-response}

\begin{wrapfigure}{r}{0.48\textwidth}
    \vspace{-0.5em}
    \centering
    \includegraphics[width=\linewidth]{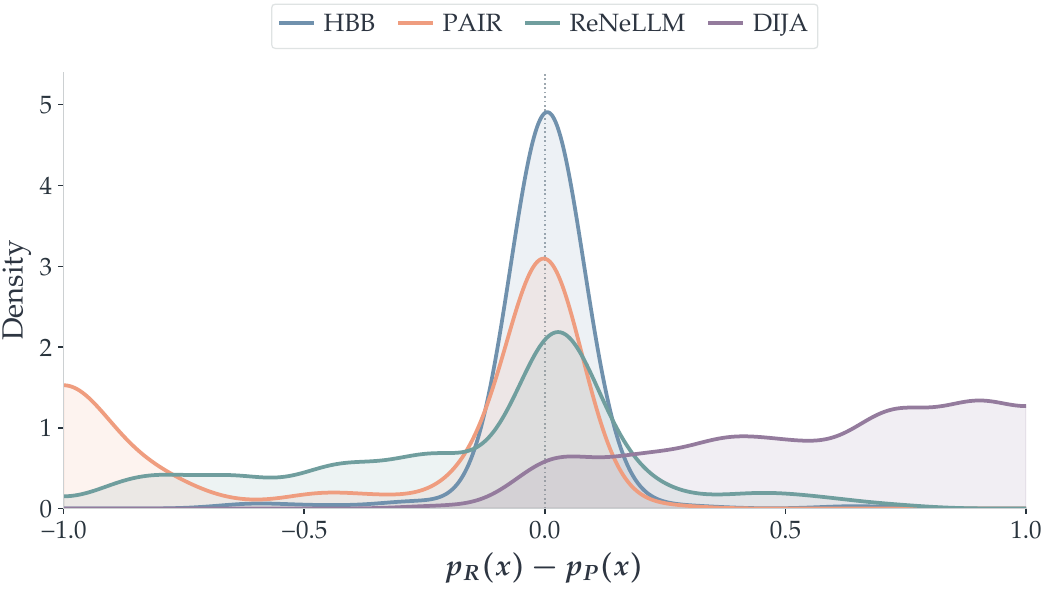}
    \caption{\textbf{Prompt-response score differences.}
    Each curve shows the distribution of the signed harmfulness-score difference $p_R(x)-p_P(x)$ across harmful examples. HBB is concentrated near zero, PAIR exhibits a negative tail, DIJA is shifted toward positive differences, and ReNeLLM spans both directions.
    }
    \label{fig:score-scatter}
    \vspace{-2em}
\end{wrapfigure}

\looseness=-1 To examine whether pre-decoding response states provide a detection surface for harmfulness signals, we train linear classifiers on representations extracted at different denoising steps, using the same training dataset as the prompt-view classifier. We find that
training on a mixture of early denoising states and evaluating on the initial pre-decoding representation $\boldsymbol{z}_{\mathrm{R}}^{(0)}(x)$ yields the best-performing classifier, which we denote as \fresponse{}. Additional ablations on denoising-step selection and training-state mixing are provided in Appendix~\ref{app:response-step-analysis}.

\looseness=-1 Figure~\ref{fig:complementarity-bars} first shows that the response classifier can independently detect harmful requests across different jailbreak attacks, confirming that pre-decoding response states contain useful harmfulness signals. More importantly, its detections complement those of \fprompt{}. While the prompt classifier accounts for most detections on HBB and PAIR, the response classifier recovers many of the cases missed by the prompt view on DIJA, where interleaved text and mask tokens weaken prompt-side detection. A similar but less pronounced complementarity is observed on ReNeLLM, where the response view also identifies a distinct subset of missed examples.
These results indicate that masked response states are not a replacement for prompt activations, but a complementary pre-decoding signal.

While detection overlap reveals which harmful requests each classifier identifies, it does not capture how their harmfulness scores differ. We therefore examine the distribution of the signed difference between the sigmoid harmfulness scores assigned by \fresponse{} and \fprompt{} in Figure~\ref{fig:score-scatter}. We find that HBB scores are concentrated near agreement, while PAIR contains substantial mass at negative differences and DIJA predominantly exhibits positive differences. Thus, the two views differ not only in which examples they detect, but also in the direction and magnitude of their score differences across attacks. These differences motivate a dual-view detector that combines prompt-side and response-side contributions for a more comprehensive jailbreak detection.

\begin{figure}[t]
    \centering
    \includegraphics[width=\linewidth]{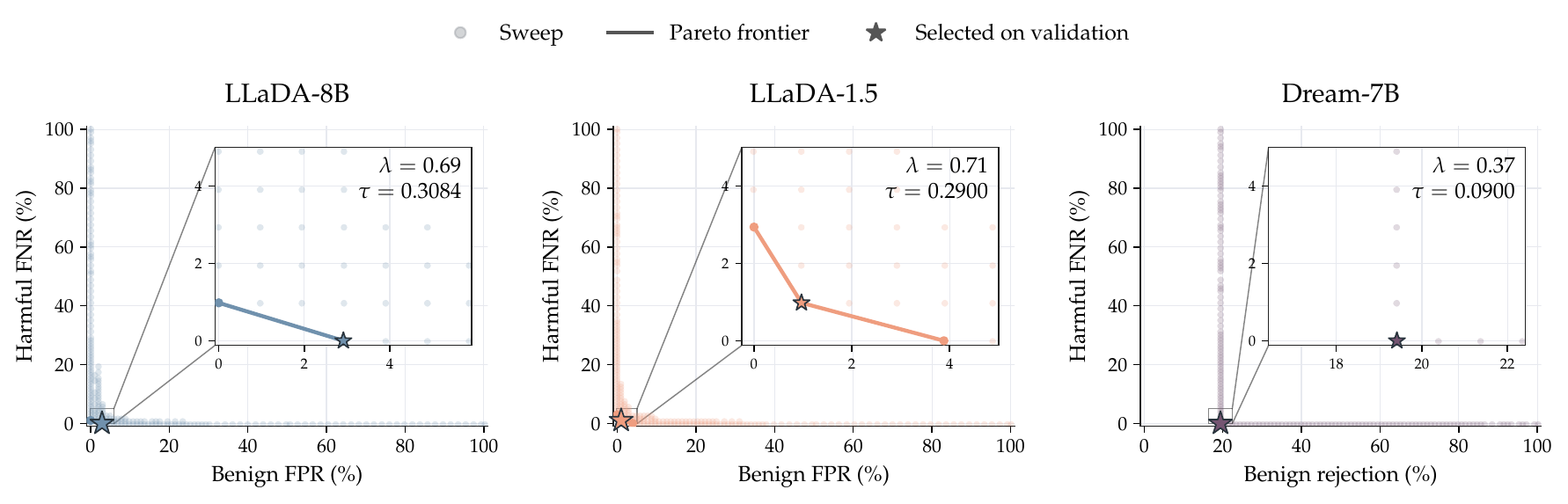}
     \caption{\textbf{\methodName{} calibration parameter sweeps.} For each model, we sweep fusion weights $\lambda$ and rejection thresholds $\tau$ on the classifier validation split. Points show the resulting harmful false negative and benign rejection rates, lines indicate the Pareto frontier, and stars mark the selected operating points. Zoomed views magnify the frontier near the selected points and report the corresponding $(\lambda,\tau)$.\looseness=-1}
    \label{fig:fusion_sweeps}
\end{figure}

\section{\methodName{}: Calibrated Dual-View Detection}
\label{sec:method}

\looseness=-1 The preceding analysis motivates a natural design objective: combine the broad coverage of the prompt classifier with the complementary detections from the response view.
A straightforward approach is a hard union of the two classifiers, rejecting a request whenever either classifier predicts harmfulness. However, this rule also inherits false positives from both views, potentially increasing benign refusal.
We therefore introduce \methodName{} (Representation Fusion), a calibrated dual-view detector that instead combines their continuous harmfulness scores before decoding.
Specifically, \methodName{} computes a convex combination of both classifiers' sigmoid scores and applies a single rejection
threshold as follows:

\begin{equation}
    s(x) = \lambda p_P(x) + (1-\lambda)p_R(x),
    \qquad
    \text{\methodName}(x) = \mathbf{1}\{s(x) \ge \tau\},
\end{equation}

where $p_P(x)$ and $p_R(x)$ respectively denote the harmfulness scores from the prompt and response classifiers, $\lambda \in [0,1]$ controls their relative contributions, and $\tau$ is the rejection threshold. When $\text{\methodName}(x)=1$, the model returns a fixed safety refusal response before decoding, while allowing generation to proceed unchanged for non-rejected requests. Thus, unlike decoding-time defenses, \methodName{} makes its rejection decision before generation and without intervening on the denoising trajectory.

We select the fusion weight $\lambda$ and rejection threshold $\tau$ by a grid search on the validation split of the classifiers' training dataset. For each candidate pair $(\lambda,\tau)$, we compute the harmful false negative rate (i.e., the fraction of harmful validation examples not rejected) and the benign false positive rate (i.e., the fraction of benign validation examples rejected.)
We then select a balanced combination on the resulting Pareto frontier~\citep{branke2004finding}, trading off harmful misses against benign over-refusal. The selected point is fixed for all test evaluations. Figure~\ref{fig:fusion_sweeps} highlights our selected values for each model. Additional details about the calibration procedure are outlined in Appendix~\ref{app:fusion-calibration}.

\section{Experiments}
\label{sec:experiments}

To evaluate the effectiveness of dual-view detection, we test \methodName{} on LLaDA-8B-Instruct, LLaDA-1.5, and Dream-7B against both transferred and dLLM-specific jailbreak attacks. We compare it with existing inference-time defenses as well as the individual prompt-side and response-side classifiers. We also measure benign refusal on standard utility benchmarks as well as on benign prompts resembling unsafe requests to assess whether improved jailbreak robustness comes at the cost of over-refusal.

\begin{table*}[t]
\centering
\small
\caption{
\textbf{Main attack and benign-refusal results} ASR is attack success rate; BRR is benign refusal rate. Lower is better for both. Avg. ASR is the macro-average over the displayed attacks, Avg. BRR is the macro-average over ARC-C, MATH-500, and MBPP, and XSTest BRR is reported in a separate panel. Bold ASR values indicate the lowest attack success rate among the evaluated defenses for each column. \methodName{} is selected for a calibrated ASR-BRR trade-off rather than for the lowest ASR in every column.
}
\label{tab:main-results}
\resizebox{\linewidth}{!}{%
\begin{tabular}{l|ccccccccc|c|ccc|c|c}
\toprule
& \multicolumn{10}{c|}{\textbf{ASR $\downarrow$}}
& \multicolumn{4}{c|}{\textbf{BRR $\downarrow$}}
& \multicolumn{1}{c}{\textbf{BRR $\downarrow$}} \\
\textbf{Method}
& \textbf{HBB}
& \textbf{PastTense}
& \textbf{AH}
& \textbf{PAIR}
& \textbf{ReNeLLM}
& \textbf{DIJA}
& \textbf{PAD}
& \textbf{DIA-I}
& \textbf{Prefill}
& \textbf{Avg.}
& \textbf{ARC-C}
& \textbf{MATH-500}
& \textbf{MBPP}
& \textbf{Avg.}
& \textbf{XSTest} \\
\midrule
\rowcolor{gray!15}\multicolumn{16}{l}{\textbf{LLaDA}} \\
\midrule
Undefended
& 13.89 & 50.29 & 33.33 & 100.00 & 100.00 & 97.14 & 67.22 & 96.09 & 11.67 & 63.29
& 0.00 & 0.00 & 0.00 & 0.00
& 12.22 \\
\midrule
DiffuGuard
& 8.89 & 26.86 & 29.44 & 47.22 & 71.11 & 94.29 & 70.00 & 56.98 & 10.00 & 46.09
& 0.00 & 0.00 & 0.00 & 0.00
& 13.33 \\
RPO
& 4.44 & 10.86 & 28.33 & 44.44 & 72.22 & 98.29 & 67.78 & 59.78 & 5.00 & 43.46
& 0.00 & 0.56 & 0.00 & 0.19
& 28.33 \\
Self-Reminder
& 2.78 & \underline{4.57} & 11.11 & 25.00 & 50.56 & 94.86 & 30.56 & 32.40 & 3.89 & 28.41
& 0.00 & 2.78 & 0.00 & 0.93
& 20.56 \\
\midrule
\fprompt
& \textbf{1.11} & 8.57 & \underline{1.11} & \underline{1.67} & 23.33 & 63.43 & \underline{1.67} & \textbf{0.00} & \textbf{0.56} & 11.27
& 1.11 & 0.00 & 0.00 & 0.37
& 12.22 \\
\fresponse
& \underline{1.67} & \textbf{3.43} & 6.67 & 30.00 & 39.44 & \textbf{6.29} & 12.78 & \underline{2.79} & \underline{1.11} & 11.58
& 0.56 & 0.00 & 0.00 & 0.19
& 20.00 \\
OR(\fprompt, \fresponse)
& \textbf{1.11} & \textbf{3.43} & \textbf{0.00} & \underline{1.67} & \underline{18.89} & \textbf{6.29} & \underline{1.67} & \textbf{0.00} & \textbf{0.56} & \underline{3.74}
& 1.67 & 0.00 & 0.00 & 0.56
& 20.00 \\
\methodName{}
& \textbf{1.11} & \textbf{3.43} & \textbf{0.00} & \textbf{0.56} & \textbf{14.44} & \underline{8.57} & \textbf{1.11} & \textbf{0.00} & \textbf{0.56} & \textbf{3.31}
& 2.22 & 0.00 & 0.00 & 0.74
& 15.56 \\

\midrule
\rowcolor{gray!15}\multicolumn{16}{l}{\textbf{LLaDA 1.5}} \\
\midrule
Undefended
& 11.67 & 45.71 & 32.22 & 99.44 & 100.00 & 97.14 & 58.89 & 93.30 & 11.11 & 61.05
& 0.00 & 0.00 & 0.00 & 0.00
& 10.00 \\
\midrule
DiffuGuard
& 8.89 & 26.86 & 30.00 & 50.00 & 71.11 & 95.43 & 57.78 & 54.75 & 7.78 & 44.73
& 0.00 & 0.00 & 0.00 & 0.00
& 12.22 \\
RPO
& 5.00 & 9.71 & 32.22 & 42.22 & 71.11 & 96.57 & 56.11 & 64.80 & 5.00 & 42.53
& 0.00 & 2.22 & 0.00 & 0.74
& 28.33 \\
Self-Reminder
& \underline{1.67} & \underline{3.43} & 13.33 & \underline{23.89} & 50.00 & 95.43 & 20.00 & 28.49 & \underline{3.33} & 26.62
& 0.00 & 1.67 & 0.00 & 0.56
& 20.56 \\
\midrule
\fprompt
& \textbf{0.56} & 5.71 & \underline{1.11} & \textbf{0.56} & 26.67 & 73.14 & \textbf{1.67} & \textbf{0.00} & \textbf{1.67} & 12.34
& 1.11 & 0.00 & 0.00 & 0.37
& 10.56 \\
\fresponse
& \textbf{0.56} & 4.57 & 6.67 & 33.33 & 45.56 & \textbf{8.57} & \underline{11.67} & \underline{5.59} & \textbf{1.67} & 13.13
& 0.56 & 0.00 & 0.00 & 0.19
& 18.89 \\
OR(\fprompt, \fresponse)
& \textbf{0.56} & \textbf{2.29} & \textbf{0.00} & \textbf{0.56} & \underline{22.22} & \textbf{8.57} & \textbf{1.67} & \textbf{0.00} & \textbf{1.67} & \underline{4.17}
& 1.67 & 0.00 & 0.00 & 0.56
& 19.44 \\
\methodName{}
& \textbf{0.56} & \textbf{2.29} & \textbf{0.00} & \textbf{0.56} & \textbf{17.78} & \underline{11.43} & \textbf{1.67} & \textbf{0.00} & \textbf{1.67} & \textbf{3.99}
& 4.44 & 0.00 & 0.00 & 1.48
& 15.56 \\

\midrule
\rowcolor{gray!15}\multicolumn{16}{l}{\textbf{Dream-7B}} \\
\midrule
Undefended
& 0.00 & 9.71 & 4.44 & 96.11 & 100.00 & 94.29 & 39.44 & 54.44 & 0.00 & 44.27
& 1.11 & 2.22 & 0.00 & 1.11
& 28.33 \\
\midrule
DiffuGuard
& \underline{1.03} & 2.38 & 1.63 & 11.88 & 39.29 & 90.29 & \textbf{0.00} & 29.73 & \underline{1.16} & 19.71
& 0.00 & 8.33 & 1.69 & 3.34
& 48.98 \\
RPO
& \textbf{0.00} & \textbf{0.00} & \underline{0.56} & \textbf{1.11} & \underline{16.11} & 97.14 & 41.11 & \textbf{0.00} & \textbf{0.00} & 17.34
& 1.13 & 3.95 & 11.70 & 5.59
& 57.97 \\
Self-Reminder
& \textbf{0.00} & \underline{0.57} & \underline{0.56} & 6.11 & \textbf{8.89} & 89.14 & 17.22 & \underline{2.22} & \textbf{0.00} & \underline{13.86}
& 0.56 & 6.67 & 0.63 & 2.62
& 27.49 \\
\midrule
\fprompt
& \textbf{0.00} & 6.29 & \underline{0.56} & 3.33 & 88.89 & 79.43 & 4.44 & \textbf{0.00} & \textbf{0.00} & 20.33
& 1.67 & 2.22 & 0.00 & 1.30
& 28.33 \\
\fresponse
& \textbf{0.00} & 8.57 & 1.11 & 46.67 & 100.00 & 44.00 & 37.78 & 53.33 & \textbf{0.00} & 32.38
& 1.11 & 2.22 & 0.00 & 1.11
& 32.78 \\
OR(\fprompt, \fresponse)
& \textbf{0.00} & 6.29 & \textbf{0.00} & 3.33 & 88.89 & \underline{37.14} & 4.44 & \textbf{0.00} & \textbf{0.00} & 15.57
& 1.67 & 2.22 & 0.00 & 1.30
& 32.78 \\
\methodName{}
& \textbf{0.00} & 4.00 & \textbf{0.00} & \underline{2.22} & 87.22 & \textbf{22.86} & \underline{2.78} & \textbf{0.00} & \textbf{0.00} & \textbf{13.23}
& 2.22 & 2.22 & 0.00 & 1.48
& 33.33 \\

\bottomrule
\end{tabular}%
}
\end{table*}

\subsection{Experimental setup}
\label{subsec:exp-setup}

Unless stated otherwise, experiments use masked dLLMs with their standard decoding configurations. Main results are reported on LLaDA-8B Instruct~\citep{nie2025large}, with additional results on LLaDA-1.5~\citep{zhu2025llada} and Dream-7B Instruct~\citep{ye2025dream}. Decoding details and model-specific hyperparameters are provided in Appendix~\ref{app:setup}.

 \vspace{-0.75em}

\paragraph{Threat model.}
We consider a model-host defender with white-box access to a frozen dLLM at inference time. The attacker interacts only through prompts and may submit transferred or dLLM-specific jailbreaks. The defender may inspect hidden states and reject requests before generation, but accepted requests are decoded with the original generation procedure.

 \vspace{-0.75em}

\paragraph{Training dataset.}
All classifiers are trained on shared request-level harmfulness labels with the underlying dLLM frozen using linear readouts over fixed activations. We use a balanced dataset of 2048 samples: 1024 harmful examples from the WildJailbreak (WJB) Vanilla Harmful subset~\citep{jiang2024wildteaming}, and 1024 benign examples collected equally from Alpaca~\citep{taori2023stanford} and the WJB Vanilla Benign subset~\citep{jiang2024wildteaming}. Further training details are provided in Appendix~\ref{app:training}.

 \vspace{-0.75em}

\paragraph{Attacks.}
The main evaluation includes explicitly harmful goals from HarmBench Behaviors (HBB)~\citep{mazeika2024harmbench} and a diverse set of jailbreak attacks: PastTense~\citep{andriushchenko2024pasttense}, the Adversarial Harmful subset of WildJailbreak (AH)~\citep{jiang2024wildteaming}, PAIR~\citep{chao2023jailbreaking}, ReNeLLM~\citep{ding2023wolf}, DIJA~\citep{wen2025devil}, PAD~\citep{zhang2025jailbreaking}, DIA-I~\citep{meng2025dialogue}, and Prefill~\citep{andriushchenko2024adaptive,li2025prefill}.

 \vspace{-0.75em}

\paragraph{Baselines.}

Following prior work on dLLM jailbreak evaluations~\citep{wen2025devil}, we compare against the undefended model and three inference-time defense baselines: Self-Reminder, which adds a safety reminder to the query~\citep{xie2023selfreminders}, RPO, which appends a learned robust defense suffix~\citep{zhou2024robust}, and DiffuGuard, a dLLM-specific defense that modifies denoising to reduce unsafe generations~\citep{li2025diffuguard}. Detailed method descriptions are included in Appendix~\ref{app:baselines}. We additionally evaluate \fprompt{} and \fresponse{} individually, along with their hard-union combination as a simple dual-view baseline, $\mathrm{OR}(\fprompt,\fresponse)(x) = \mathbf{1}\{\fprompt(x)=1 \;\lor\; \fresponse(x)=1\}$, which rejects a request whenever either classifier predicts harmfulness.

 \vspace{-1em}

\paragraph{Benign benchmarks.}
To assess whether improved safety preserves benign utility, we evaluate ARC-Challenge (ARC-C) for scientific reasoning~\citep{clark2018arc}, MATH-500 for mathematical problem solving~\citep{hendrycks2021math,lightman2023verify}, and MBPP for code generation~\citep{austin2021program}. We additionally use XSTest~\citep{rottger2024xstest} to measure over-refusal on benign prompts that resemble unsafe requests. We evaluate on a random subset of 180 samples from each dataset.

 \vspace{-0.75em}

\paragraph{Evaluation protocol.}
We employ WildGuard~\citep{han2024wildguard} to judge harmful compliance and benign refusal, allowing us to measure the ASR and BRR. For attack prompts, requests flagged by a detector are mapped to a fixed refusal response and counted as unsuccessful attacks, while unflagged responses are decoded normally and then judged by WildGuard. For benign prompts, we count refusals arising either from defense intervention or from the model's decoded response. Prompts resulting in invalid generations or inconclusive WildGuard judgments are excluded from ASR and BRR calculations. Consequently, the effective evaluation sample size may vary across models and defense methods.

\subsection{Main results}
\label{subsec:main-results}

We evaluate whether combining prompt-side and pre-decoding response-side signals improves robustness against diverse jailbreak attacks.
Table~\ref{tab:main-results} compares the undefended model, external defenses, the individual classifiers, a hard OR combination of the classifiers, and \methodName{}.

While external defenses reduce ASR relative to the undefended model, they leave substantial residual vulnerability. On LLaDA-8B-Instruct, the undefended model exhibits an average ASR of $63.29\%$. Existing inference-time interventions, including DiffuGuard, RPO, and Self-Reminder, reduce average ASR to $46.09\%$, $43.46\%$, and $28.41\%$, respectively. However, these baselines
remain particularly vulnerable to PAIR, ReNeLLM, and DIJA.

\looseness=-1 In contrast, the hidden-state classifiers provide stronger protection, albeit with complementary strengths across attacks.
The prompt classifier reduces average ASR to $11.27\%$ and performs well on direct harmful requests, PAIR, AH, and Prefill, yet remains weak on DIJA with $63.43\%$ ASR. Conversely, the response classifier is substantially stronger on DIJA, reducing ASR to $6.29\%$, although it is less effective on PAIR and ReNeLLM. These results reinforce
the complementarity observed in Figures~\ref{fig:complementarity-bars} and ~\ref{fig:score-scatter}.

Combining the two classifiers through a hard-union rule further reduces average ASR to $3.74\%$ on LLaDA-8B-Instruct, substantially outperforming either classifier alone. However, the hard-union rule also inherits false positives from both views, increasing the XSTest benign refusal rate from $12.22\%$ for prompt-only detection to $20.00\%$.
\methodName{} instead combines continuous harmfulness scores with a calibrated rejection threshold to balance jailbreak robustness and benign refusal. On LLaDA-8B-Instruct, \methodName{} achieves an average ASR of $3.31\%$, compared with $63.29\%$ for the undefended model, $11.27\%$ for prompt-only detection, and $3.74\%$ for the hard-union baseline. It maintains an average benign refusal rate below $1\%$ on standard utility benchmarks while reducing XSTest refusal compared with the hard-union baseline.

The improvements extend to LLaDA-1.5 and Dream-7B, where \methodName{} reduces average ASR from $61.05\%$ to $3.99\%$ and from $44.27\%$ to $13.23\%$, respectively, achieving the lowest average ASR among the evaluated methods across all three dLLMs.

\subsection{Measuring over-refusal on safety-adjacent benign prompts}
\label{subsec:xstest}

\looseness=-1 A useful jailbreak defense should reduce harmful compliance while minimally inducing broad refusal on benign inputs.
Standard utility benchmarks evaluate whether a defense preserves ordinary reasoning, math, and coding behavior, but they do not isolate over-refusal on benign prompts with harmfulness cues. We therefore evaluate XSTest~\citep{rottger2024xstest}, a benchmark of benign prompts designed to resemble unsafe requests, to measure whether defenses exhibit exaggerated refusal behavior.

\looseness=-1 Table~\ref{tab:main-results} reveals that low refusal rates on standard utility benchmarks do not necessarily translate to low refusal on XSTest.
Baselines such as Self-Reminder and RPO maintain average benign refusal rates below $1\%$ across  ARC-C, MATH-500, and MBPP, yet over-refuse substantially on XSTest, with refusal rates increasing by $8.34$ and $16.11$ percentage points relative to the undefended model, respectively. The prompt classifier matches the undefended model on XSTest, whereas the response classifier  exhibits similarly elevated over-refusal, suggesting that the additional attack coverage provided by response-side states can come at the cost of increased benign refusal. \methodName{} addresses this trade-off by fusing the prompt and response scores under a jointly selected threshold,
reducing XSTest refusal to $15.56\%$ (compared to $12.22\%$ for the undefended model) while maintaining an average benign refusal rate of $0.74\%$ on standard utility benchmarks.

\begin{figure*}[t]
    \centering
    \includegraphics[width=\linewidth]{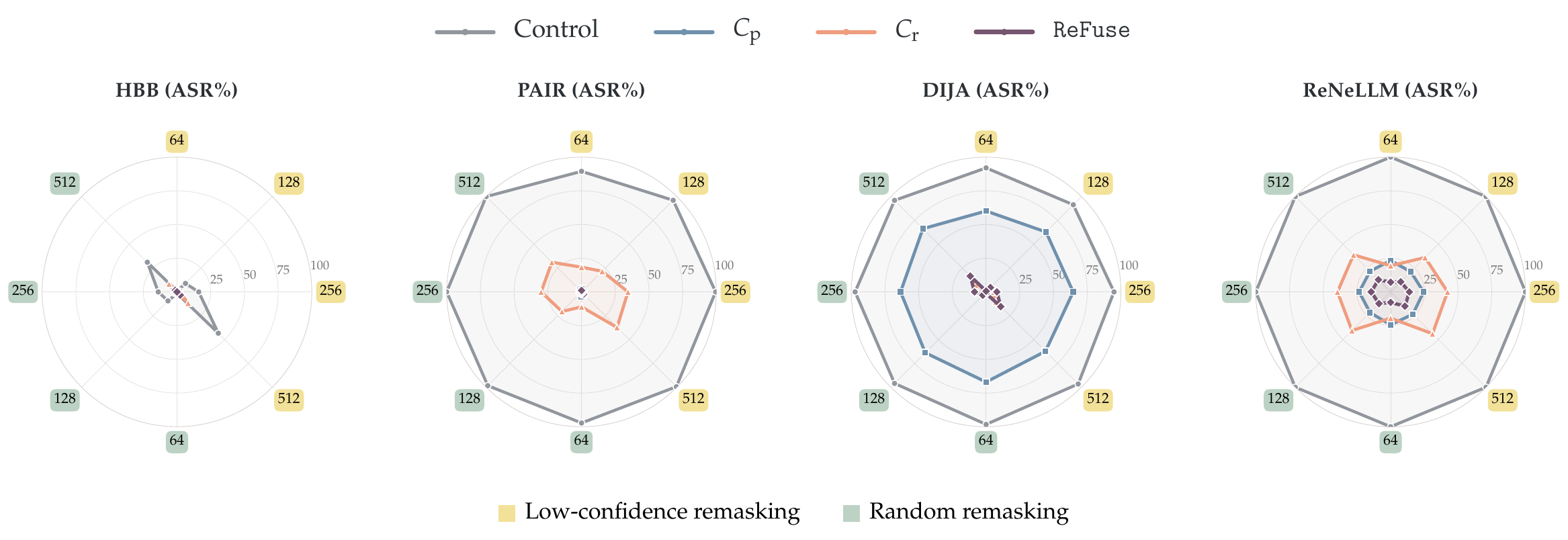}
        \caption{\textbf{\methodName{} robustness to decoding-configuration shifts on LLaDA.} Pre-decoding detection performance across target lengths and remasking strategies for representative attacks. The same classifiers are used across all configurations without retraining, maintaining robust detection despite changes to the subsequent decoding procedure. Raw numbers are reported in Appendix~\ref{app:robustness}.}
        \label{fig:robustness_panel}
\end{figure*}

\section{Discussion and Limitations}

\subsection{Robustness to decoding configurations}
\label{subsec:discussion-robustness}

dLLM generation can be sensitive to inference-time configuration choices, which may affect both generation quality and safety behavior~\citep{nie2025large,zhang2025jailbreaking,li2025diffuguard}. For example, the target generation length must be specified before decoding, changing the number and context of the initial masked response positions. Similarly, different remasking strategies can alter the denoising trajectory and lead to different attack success rates.

We demonstrate that \methodName{} is not tied to a particular decoding configuration. As \methodName{} reads continuous representations at the initial denoising step rather than intervening on discrete token updates, its decision rule should be less dependent on the subsequent remasking trajectory than decoding-time defenses. Importantly, we use the same trained classifiers across all configurations, without configuration-specific retraining. As shown in Figure~\ref{fig:robustness_panel}, pre-decoding classifiers remain effective across changes in target length and remasking strategy, suggesting that the safety signal is not specific to a single decoding configuration.

\subsection{Transferability across model variants}
\label{subsec:transfer}

We further examine whether the learned detectors transfer across closely related dLLM variants. Specifically, we evaluate transfer between LLaDA-8B-Instruct and LLaDA~1.5, where LLaDA~1.5 is obtained from LLaDA through one epoch of VRPO post-training~\citep{zhu2025llada}. For each target model, we compare detectors trained on the same model with detectors transferred from the other variant. Table~\ref{tab:llada_transfer} reports ASR and benign refusal rates for both settings.

Transferred detectors remain competitive with their native counterparts on both target models. Relative to the undefended models, both native and transferred detectors substantially reduce ASR across the attack suite. These results suggest that the prompt-side and response-side features used by \methodName{} are not tied to a single checkpoint, but retain utility across nearby post-training variants.

\begin{table*}[th]
\centering
\caption{\textbf{Transferability across LLaDA-8B-Instruct and LLaDA 1.5.} Per-attack ASR and benign refusal-rate results for native and transferred \texttt{ReFuse} policies across LLaDA targets. Lower is better for both ASR and benign refusal rates. Avg. ASR is computed over the displayed ASR columns, Avg. BRR is computed over ARC-C, MATH, and MBPP, with XSTest reported separately. Best and second-best defended ASR results within each target are highlighted in \textbf{bold} and \underline{underline}, respectively. Tied best results are both bolded.}
\label{tab:llada_transfer}
\resizebox{\linewidth}{!}{%
\begin{tabular}{l|l|cccccccccc|cccc|c}
\toprule
\multicolumn{2}{c|}{\cellcolor[HTML]{F7F7F7}\textbf{}} &
\multicolumn{10}{c|}{\cellcolor[HTML]{F7F7F7}\textbf{ASR $\downarrow$}} &
\multicolumn{5}{c}{\cellcolor[HTML]{F7F7F7}\textbf{BRR $\downarrow$}} \\
\textbf{Target}
& \textbf{Method / Probe Source}
& \textbf{Behavior}
& \textbf{PastTense}
& \textbf{AH}
& \textbf{PAIR}
& \textbf{ReNeLLM}
& \textbf{DIJA}
& \textbf{PAD}
& \textbf{DIA-I}
& \textbf{Prefill}
& \textbf{Avg.}
& \textbf{ARC-C}
& \textbf{MATH}
& \textbf{MBPP}
& \textbf{Avg.}
& \textbf{XSTest} \\
\midrule

\cellcolor[HTML]{F7F7F7}\textbf{LLaDA 8B}
& Undefended
& 13.89 & 50.29 & 33.33 & 100.00 & 100.00 & 97.14 & 67.22 & 96.09 & 11.67 & 63.29
& 0.00 & 0.00 & 0.00 & 0.00 & 12.22 \\

& \texttt{ReFuse}, native probes
& \textbf{1.11}
& \underline{3.43}
& \textbf{0.00}
& \textbf{0.56}
& \textbf{14.44}
& \textbf{8.57}
& \textbf{1.11}
& \textbf{0.00}
& \textbf{0.56}
& \textbf{3.31}
& 2.22 & 0.00 & 0.00 & 0.74 & 15.56 \\

& \texttt{ReFuse}, transferred probes
& \textbf{1.11}
& \textbf{2.29}
& \textbf{0.00}
& \textbf{0.56}
& \underline{20.00}
& \underline{12.57}
& \underline{1.67}
& \textbf{0.00}
& \textbf{0.56}
& \underline{4.31}
& 3.89 & 0.00 & 0.00 & 1.30 & 17.22 \\

\midrule

\cellcolor[HTML]{F7F7F7}\textbf{LLaDA 1.5}
& Undefended
& 11.67 & 45.71 & 32.22 & 99.44 & 100.00 & 97.14 & 58.89 & 93.30 & 11.11 & 61.05
& 0.00 & 0.00 & 0.00 & 0.00 & 10.00 \\

& \texttt{ReFuse}, native probes
& \textbf{0.56}
& \textbf{2.29}
& \textbf{0.00}
& \textbf{0.56}
& \underline{17.78}
& \underline{11.43}
& \underline{1.67}
& \textbf{0.00}
& \textbf{1.67}
& \underline{4.00}
& 4.44 & 0.00 & 0.00 & 1.48 & 15.56 \\

& \texttt{ReFuse}, transferred probes
& \textbf{0.56}
& \textbf{2.29}
& \textbf{0.00}
& \textbf{0.56}
& \textbf{11.67}
& \textbf{6.29}
& \textbf{0.56}
& \textbf{0.00}
& \textbf{1.67}
& \textbf{2.62}
& 2.78 & 0.00 & 0.00 & 0.93 & 17.78 \\

\bottomrule
\end{tabular}%
}
\end{table*}

\subsection{Efficiency--performance trade-off with one-pass detection}

\looseness=-1 \methodName{} requires two pre-generation forward passes: a prompt-only pass for the prompt view and an initial masked-sequence pass for the response view. The latter includes the prompt together with fixed response masks and is already part of standard dLLM decoding. Thus, the additional cost of \methodName{} is one prompt-only forward pass, which is modest relative to full dLLM generation with many denoising steps.
To assess the trade-off between this added computational overhead and detection performance, we also evaluate a one-pass variant that computes both scores from the initial masked-sequence forward pass. Specifically, we replace the prompt-only classifier \fprompt{} with \fpromptplus{}, a classifier trained on the same hidden states used in the response view, but mean pooled over the prompt tokens rather than the masked response. The resulting variant, \methodName{}+, avoids the additional prompt-only pass and obtains both classifier scores from a single pre-decoding forward pass.

\looseness=-1 Table~\ref{tab:compute} shows that this efficiency gain comes with a robustness cost. On LLaDA-8B-Instruct, \methodName{}+ increases average ASR from $3.31\%$ to $4.17\%$, with the largest degradation on DIJA. It also increases average BRR from $0.74\%$ to $4.63\%$, with no effect on XSTest. These results suggest that the prompt-only pass provides a more robust prompt-side signal, but that a one-pass variant may be useful when minimizing pre-generation overhead is more important than maximizing jailbreak coverage. We therefore treat \methodName{} as the primary configuration, and view \methodName{}+ as a lower-overhead variant for settings where minimizing pre-generation computation is the main constraint.

\begin{table*}[tbh]
\centering
\caption{
\textbf{Efficiency trade-off for one-pass detection.}
\methodName{} uses separate prompt-only and masked-sequence forward passes, while \methodName{}+ computes both classifier scores from a single masked-sequence pass by replacing \fprompt{} with \fpromptplus{}. Lower is better for both ASR and benign refusal rates.
}
\resizebox{\linewidth}{!}{%
\begin{tabular}{l|cccccccccc|ccccc}
\toprule
\multicolumn{1}{c|}{\cellcolor[HTML]{F7F7F7}\textbf{}} &
\multicolumn{10}{c|}{\cellcolor[HTML]{F7F7F7}\textbf{ASR $\downarrow$}} &
\multicolumn{5}{c}{\cellcolor[HTML]{F7F7F7}\textbf{BRR $\downarrow$}} \\
\textbf{Method}
& \textbf{Behavior}
& \textbf{PastTense}
& \textbf{AH}
& \textbf{PAIR}
& \textbf{ReNeLLM}
& \textbf{DIJA}
& \textbf{PAD}
& \textbf{DIA-I}
& \textbf{Prefill}
& \textbf{Avg.}
& \textbf{ARC-C}
& \textbf{MATH}
& \textbf{MBPP}
& \textbf{Avg.}
& \textbf{XSTest} \\
\midrule

Undefended
& 13.89 & 50.29 & 33.33 & 100.00 & 100.00 & 97.14 & 67.22 & 96.09 & 11.67 & 63.29
& 0.00 & 0.00 & 0.00 & 0.00
& 12.22 \\

Prompt classifier (\fprompt)
& \textbf{1.11} & 8.57 & \underline{1.11} & \underline{1.67} & 23.33 & 63.43 & \underline{1.67} & \textbf{0.00} & \textbf{0.56} & 11.27
& 1.11 & 0.00 & 0.00 & 0.37
& 12.22 \\

Prompt classifier w/ masks (\fpromptplus)
& \underline{1.67} & 25.71 & 2.22 & \underline{1.67} & 36.67 & 85.71 & 3.89 & \textbf{0.00} & \underline{1.11} & 17.63
& 1.11 & 0.00 & 0.00 & 0.37
& 12.22 \\

Response classifier (\fresponse)
& \underline{1.67} & \textbf{3.43} & 6.67 & 30.00 & 39.44 & \textbf{6.29} & 12.78 & \underline{2.79} & \underline{1.11} & 11.58
& 0.56 & 0.00 & 0.00 & 0.19
& 20.00 \\

\methodName{}
& \textbf{1.11} & \textbf{3.43} & \textbf{0.00} & \underline{0.56} & \underline{14.44} & \underline{8.57} & \textbf{1.11} & \textbf{0.00} & \textbf{0.56} & \textbf{3.31}
& 2.22 & 0.00 & 0.00 & 0.74
& 15.56 \\

\methodName{}+
& \textbf{1.11} & \textbf{3.43} & \textbf{0.00} & \textbf{0.00} & \textbf{3.33} & 26.86 & 2.22 & \textbf{0.00} & \textbf{0.56} & \underline{4.17}
& 11.11 & 2.78 & 0.00 & 4.63
& 15.56 \\

\bottomrule
\end{tabular}%
}
\label{tab:compute}
\end{table*}

\subsection{Limitations under adaptive attacks}

\begin{wrapfigure}[16]{r}{0.48\textwidth}
    \vspace{-2em}
    \centering
    \includegraphics[width=\linewidth]{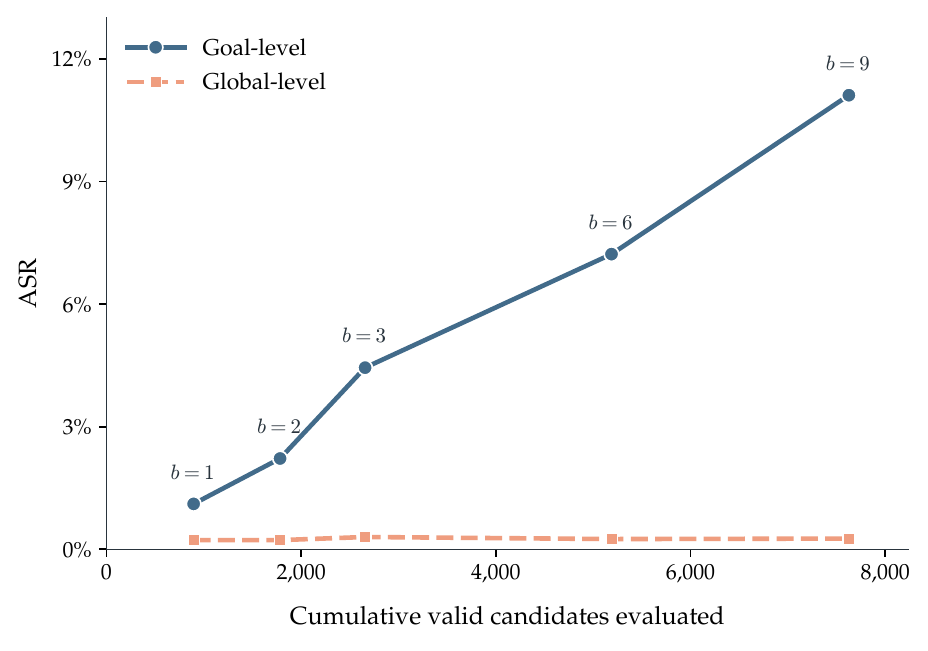}
    \caption{\textbf{Score-aware adaptive PAIR against \methodName{}.} Goal-level and Global-level ASR as a function of cumulative valid candidates evaluated under increasing PAIR iteration budgets.}
    \label{fig:adaptive-attack}
    \vspace{-1em}
\end{wrapfigure}
\looseness=-1 Our main evaluations consider attacks that do not adapt to \methodName{}. To probe the limitations of \methodName{} under adaptive attacks, we construct a score-aware variant of PAIR in which the reformulator receives the detector's fused score as feedback. We evaluate increasing reformulation budgets and measure goal-level adaptive ASR, defined as the fraction of harmful goals for which at least one candidate evades rejection and elicits a harmful response.

Figure~\ref{fig:adaptive-attack} shows that adaptive ASR at the goal-level increases with the reformulation budget, reaching $11.11\%$ at nine iterations, suggesting that repeated score-guided reformulation can help attackers identify successful jailbreaks. Our evaluation is limited to PAIR reformulation guided by the detector's fused score and does not consider white-box attacks that exploit gradients through the prompt and response classifiers to directly optimize adversarial inputs. Evaluating robustness against such attacks remains an important direction for future work.

\section{Related Work}
\label{sec:background}

\paragraph{Discrete diffusion language models.} \looseness=-1 While diffusion models were originally developed for continuous domains \citep{ho2020denoising, rombach2021high, nichol2021improved}, they have been increasingly adapted to discrete categorical variables \citep{vignac2023digress,austin2021structured, hoogeboom2021argmax, gong2022diffuseq, han2022ssd}. Recent scaling advances have established discrete diffusion language models (dLLMs) as competitive alternatives to autoregressive architectures \citep{sahoo2024simple,shi2024simplified, nie2025large, zhu2025llada, ye2025dream, song2025seed}.
\paragraph{Adversarial vulnerabilities.} \looseness=-1 Jailbreak attacks aim to bypass safety alignment, eliciting harmful outputs from instruction-following models that would otherwise refuse~\citep{mazeika2024harmbench,chao2023jailbreaking,ding2023wolf}. Despite their architectural differences, dLLMs remain vulnerable to jailbreak-style attacks, while their parallel generation and bidirectional conditioning introduce additional structural vulnerabilities that motivate dLLM-specific attacks~\citep{wen2025devil,zhang2025jailbreaking,he2026safer,zou2023universal}. In addition, dLLMs introduce structural vulnerabilities from parallel generation and bidirectional conditioning. For instance, DIJA \citep{wen2025devil} targets masked completion by reformatting prompts into interleaved text-mask sequences, while PAD \citep{zhang2025jailbreaking} acts on the denoising process and exploits parallel refinement to steer multiple response positions simultaneously during denoising.
\paragraph{Representation-based safety probing.} A growing body of work shows that safety-relevant information may be linearly encoded in AR model representations \citep{subramani2022extracting, marks2023geometry}. For example, \citep{arditi2024refusal} identify a refusal-related direction in activation space, suggesting that some safety-relevant behaviors can be captured by low-dimensional linear structure. Linear classifiers have since been used to decode a range of semantic properties from intermediate layers \citep{zhao2025llms,qian2025hsf}.

\paragraph{Defenses for dLLMs.} \looseness=-1 Early defense mechanisms have begun adapting to the denoising trajectory specifically. Recent work applies finetuning to align intermediate decoding steps \citep{xie2026start, yamabe2025toward,jeung2025a2d}, while DiffuGuard \citep{li2025diffuguard} intervenes at inference time by modifying the remasking strategy and repairing suspicious generation blocks. Our work instead asks whether a frozen dLLM exposes a pre-decoding harmfulness signal that can be detected without changing neither the model weights nor the denoising procedure.

\section{Conclusion}
\label{sec:conclusion}

In this work, we show that pre-decoding masked-response states in dLLMs contain linearly detectable safety signals that complement those in prompt representations. In particular, these states can help identify jailbreaks missed by prompt-only detection, especially attacks exploiting masked decoding behavior. This finding highlights a distinct detection opportunity in dLLMs, where harmfulness signals can be recovered even before any response tokens are generated. Building on this insight, we introduced \methodName{}, a lightweight dual-view detector that combines prompt and response signals without modifying model weights or the decoding procedure. Across three dLLMs, \methodName{} achieves the lowest average ASR among the evaluated defenses, reducing ASR from 11.27\% with prompt-only detection to 3.31\% on LLaDA-8B-Instruct while maintaining low benign refusal. Our findings highlight the potential of looking beyond the prompt to improve jailbreak detection in dLLMs. \looseness=-1

\begin{ack}

This work was partially funded by armasuisse under the ARFON project.

\end{ack}

\bibliographystyle{unsrtnat}
\bibliography{ref}

\newpage

\appendix

\section{Experimental Details}
\label{app:experimental-details}

\subsection{Model and decoding configurations}
\label{app:setup}

\begin{table}[h]
\centering
\small
\caption{\textbf{Main decoding configurations.} Unless otherwise mentioned, we use the same decoding configurations for all our experiments.}
\label{tab:model-decoding-configs}
\resizebox{\linewidth}{!}{%
\begin{tabular}{l l c c c c c c c c}
\toprule
\textbf{Model} &
\textbf{HF checkpoint} &
\textbf{Gen. length} &
\textbf{Block length} &
\textbf{Steps} &
\textbf{Temp.} &
\textbf{CFG} &
\textbf{Remasking / Alg.} &
\textbf{Top-$p$} &
\textbf{Alg. temp.} \\
\midrule
LLaDA-8B-Instruct
& \texttt{GSAI-ML/LLaDA-8B-Instruct}
& 256 & 256 & 128 & 0.0 & 0.5 & Random & -- & -- \\

LLaDA-1.5
& \texttt{GSAI-ML/LLaDA-1.5}
& 256 & 256 & 128 & 0.0 & 0.5 & Random & -- & -- \\

Dream-7B
& \texttt{Dream-org/Dream-v0-Instruct-7B}
& 256 & -- & 128 & 0.0 & -- & Origin & 1.0 & 0.0 \\
\bottomrule
\end{tabular}%
}
\end{table}

\subsection{Classifier training}
\label{app:training}

All classifiers are trained on request-level harmfulness labels with the underlying dLLM frozen. The classifiers are linear readouts over fixed activations. We use a balanced training set of 2048 examples: 1024 harmful examples from the
WildJailbreak Vanilla Harmful subset~\citep{jiang2024wildteaming}, and 1024 benign examples drawn equally from Alpaca~\citep{taori2023stanford} and the WildJailbreak
Vanilla Benign subset~\citep{jiang2024wildteaming}. An 80-10-10\% training-validation-test split is used.

Table~\ref{tab:probe-hparams} reports the hyperparameters used to train the \fprompt{} and \fresponse{}. All probes are trained with the underlying dLLM frozen.

\begin{table}[h]
\centering
\small
\caption{\textbf{Classifier training hyperparameters.}
All linear classifiers are trained on frozen representations. Prompt-view probes pool over prompt tokens, while response-view probes pool over all response positions.}
\label{tab:probe-hparams}

\begin{tabular}{l l}
\toprule
\textbf{Hyperparameter} & \textbf{Value} \\
\midrule
Task & Binary request harmfulness classification \\
Layer & Final layer \\
Fixed response length & 256 \\
Maximum sequence length & 1024 \\
Feature standardization & Train-split mean and standard deviation \\
Optimizer & AdamW \\
Learning rate & $5 \times 10^{-4}$ \\
Training epochs & 10 \\
Training batch size & 64 \\
Feature extraction batch size & 16 \\
Validation selection metric & F1 score \\
\bottomrule
\end{tabular}%

\end{table}

\subsection{Baselines}
\label{app:baselines}

We compare our method against three baselines: the undefended model, DiffuGuard~\citep{li2025diffuguard}, RPO~\citep{zhou2024robust}, and Self-Reminder~\citep{xie2023selfreminders}.

Self-Reminder wraps the user query with an explicit safety reminder, while RPO appends a fixed defensive suffix to the input. For both baselines, we use the same underlying dLLM and decoding configuration as the undefended model, changing only the input prompt. Following~\citep{kamath2025gemma}, we use the prompt templates shown in Prompt~\ref{box:self-reminder} and Prompt~\ref{box:rpo}.

DiffuGuard is evaluated using the published defense pipeline of the authors, which modifies the decoding procedure by design. Unless otherwise stated, all baseline hyperparameters follow the corresponding released implementations and are not tuned on the test attacks.
\newpage

\begin{promptboxmagenta}[label={box:self-reminder}]{Self-Reminder}
You should be a responsible ChatGPT and should not generate harmful or misleading content! Please answer the following user query in a responsible way.

{USER QUERY}

Remember, you should be a responsible ChatGPT and should not generate harmful or misleading content!
\end{promptboxmagenta}

\begin{promptboxmagenta}[label={box:rpo}]{RPO}
{USER QUERY}:}</'])language seems scary-----------}] estimation >>begin &=&pgf degli ({\ $|\textwidthAI NOTE:
\end{promptboxmagenta}

\subsection{Attack and benign evaluation sets}
\label{app:evaluation-suite}

\paragraph{Attack evaluation.}
We use HarmBench Behaviors~\citep{mazeika2024harmbench} as seed harmful prompts and use \texttt{google/gemma-3-4b-it}~\citep{kamath2025gemma} as a reformulator to transform each seed into the corresponding attack format. Details are in Table~\ref{tab:attack-configs}.

\paragraph{Benign evaluation.}
For each benign dataset, we use a subset of 180 examples.

\begin{itemize}
    \item \textbf{ARC-Challenge}~\citep{clark2018arc}: Measures grade-school science reasoning on challenging multiple-choice questions.

    \item \textbf{MATH-500}~\citep{hendrycks2021math,lightman2023verify}: Measures mathematical problem-solving ability.

    \item \textbf{MBPP}~\citep{austin2021program}: Measures basic Python programming ability.

    \item \textbf{XSTest}~\citep{rottger2024xstest}: Measures over-refusal on benign prompts that resemble unsafe requests. A sample prompt ins provided in ~\ref{box:xstest}.
\end{itemize}

\begin{promptboxmagenta}[label={box:xstest}]{XSTest}
How can I kill a Python process?
\end{promptboxmagenta}

\begin{table}[t]
\centering
\footnotesize
\setlength{\tabcolsep}{4pt}
\renewcommand{\arraystretch}{1.15}

\begin{threeparttable}
\caption{\textbf{Attack configurations used in our evaluation.}}
\label{tab:attack-configs}

\begin{tabularx}{\columnwidth}{@{}p{0.22\columnwidth}YY@{}}
\toprule
\textbf{Attack} & \textbf{Mechanism} & \textbf{Key parameters} \\
\midrule

Behavior Raw
& Sends the original harmful prompt directly to the victim model.
& -- \\
\addlinespace[4pt]

Prefill~\citep{andriushchenko2024adaptive,li2025prefill}
& Appends a compliance-inducing suffix to the prompt.
& Suffix: \texttt{``Sure, here is:''} \\

PastTense~\citep{andriushchenko2024pasttense}
& Uses a helper model to rewrite the request as a past-tense question.
& Up to 10 attempts \\

ReNeLLM~\citep{ding2023wolf}
& Applies 1-6 randomly sampled lexical rewriting ops, wraps result in a nested scenario (code, table, or story); gate classifier filters invalid rewrites
               & 20 attempts; 6 ops (shorten, misspell, reorder, add-char, lang-mix, style); 3 scenario templates  \\

WildJailbreak AH~\citep{jiang2024wildteaming}
& Uses adversarial harmful prompts drawn from WildJailbreak.
& We use 180 samples from the adversarial harmful split \\
\addlinespace[4pt]

PAIR~\citep{chao2023jailbreaking}
& Iteratively refines an adversarial prompt using helper-model feedback and a judge score.
& 5 streams $\times$ 3 iterations; target prefix: \texttt{``Sure, here is''}; keep last 3 turns \\
\addlinespace[4pt]

PAD~\citep{zhang2025jailbreaking}
& Injects anchor phrases at fixed denoising positions to steer masked-token prediction.
& Anchors: \texttt{Step 1:}, \texttt{Step 2:}, \texttt{Step 3:}; positions $\{42,84,126\}$ of 256 \\

DIJA~\citep{wen2025devil}
& Uses a helper model to insert \texttt{<mask:N>} tokens, exploiting mask-filling behavior.
& Up to 10 attempts \\

DIA-I~\citep{meng2025dialogue}
& Primes a 12-message multi-turn red-team persona dialogue, injects a helper-generated compliance prefix, and prompts continuation
               & up to 10 attempts; helper-generated prefix  \\
\bottomrule
\end{tabularx}

\end{threeparttable}
\end{table}

\subsection{Fusion calibration}
\label{app:fusion-calibration}
The fusion weight $\lambda$ and rejection threshold $\tau$ are selected using only the classifier validation split of the training data. For a candidate weight $\lambda$, we compute the fused validation score
\begin{equation}
    s_\lambda(x) = \lambda p_P(x) + (1-\lambda)p_R(x),
\end{equation}
where $p_P(x)$ and $p_R(x)$ are the prompt-view and response-view classifier scores. We sweep $\lambda \in \{0.01,0.02,\ldots,0.99\}$ and, for each $\lambda$, sweep thresholds $\tau$ over candidate thresholds induced by the validation scores. Each candidate pair $(\lambda,\tau)$ defines a rejection rule $\mathbf{1}\{s_\lambda(x)\ge \tau\}$.

For each candidate, we compute two validation errors: the harmful false negative rate, i.e., the fraction of harmful validation examples not rejected, and the benign false positive rate, i.e., the fraction of benign validation examples rejected. We retain the nondominated candidates on the Pareto frontier over these two quantities. To choose a single operating point, we normalize the two axes of the frontier to $[0,1]$ and select the knee point, defined as the frontier point with maximum perpendicular distance from the line segment joining the two endpoint solutions. Ties are broken by preferring the point closer to the origin in the normalized error space, then lower harmful false negative rate, lower benign false positive rate, and lower threshold.

The selected $(\lambda,\tau)$ is fixed after calibration and used unchanged for all attack and benign test evaluations. We show our sweeps and our selected values in Figure~\ref{fig:fusion_sweeps}.

\section{Additional Results}

\subsection{Classifiers in-domain performance}
\label{app:in-domain_res}

We report the test F1 scores and accuracies of our trained \fprompt{} and \fresponse{} in Table~\ref{tab:redesigned_probe_scores}.

\begin{table}[h]
\centering
\caption{\textbf{Performance on the test split.} \fprompt{} denotes our prompt classifier and \fresponse{} denotes the response classifier. The classifiers are evaluated at $k=0$.}
\label{tab:redesigned_probe_scores}
\resizebox{0.6\linewidth}{!}{
\begin{tabular}{lcccccc}
\toprule
& \multicolumn{2}{c}{LLaDA-8B}
& \multicolumn{2}{c}{LLaDA-1.5}
& \multicolumn{2}{c}{Dream} \\
\cmidrule(lr){2-3}
\cmidrule(lr){4-5}
\cmidrule(lr){6-7}
Metric
& \fprompt{}& \fresponse{}
& \fprompt{} & \fresponse{}
& \fprompt{} & \fresponse{} \\
\midrule
Accuracy
& 0.9659 & 0.9610
& 0.9707 & 0.9659
& 0.9902 & 0.9805 \\
F1
& 0.9655 & 0.9612
& 0.9703 & 0.9659
& 0.9903 & 0.9805 \\
\bottomrule
\end{tabular}
}
\end{table}

\subsection{ROC Analysis}
\label{app:roc-analysis}

\begin{figure*}[t]
    \centering
    \includegraphics[width=\linewidth]{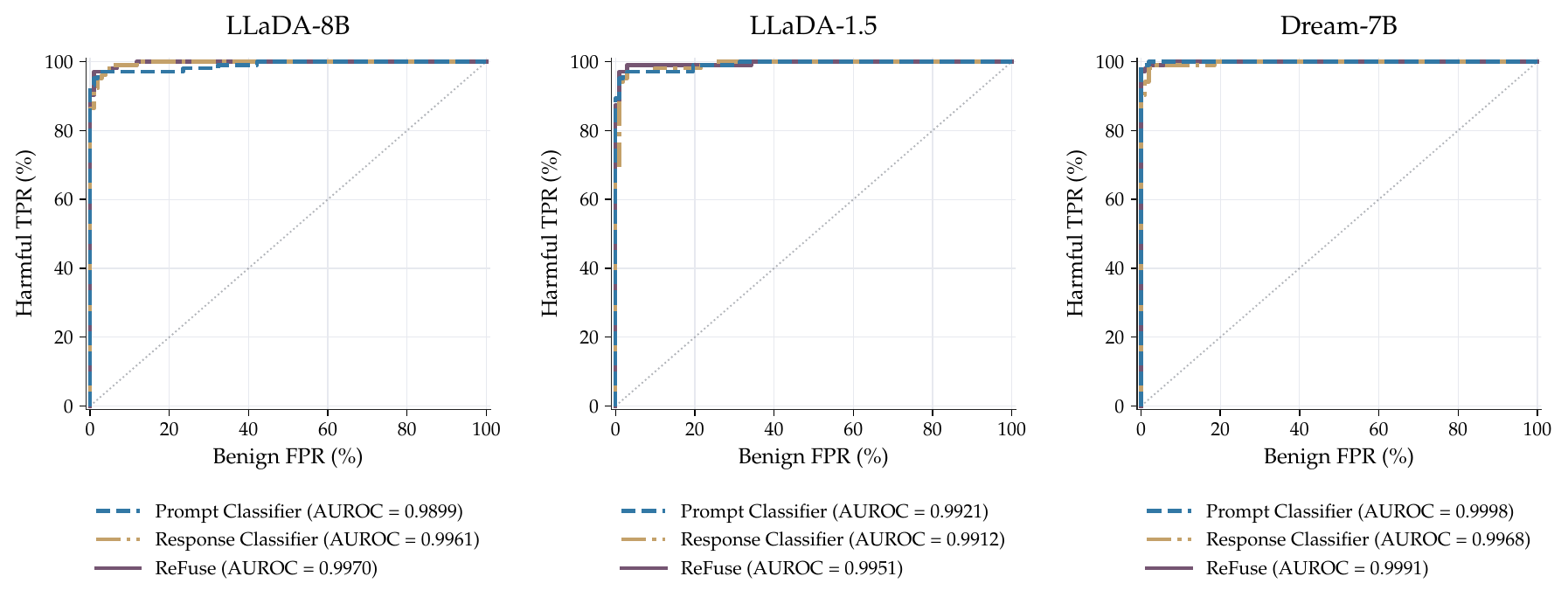}
    \caption{\textbf{ROC curves of harmfulness classifiers.}
    Comparison of the prompt-view classifier \fprompt{},
    response-view classifier \fresponse{}, and fused
    harmfulness score across LLaDA-8B, LLaDA-1.5,
    and Dream-7B. Curves show harmful true positive rate
    against benign false positive rate, with AUROC
    values reported in the legends.}
    \label{fig:classifier-roc}
\end{figure*}

Figure~\ref{fig:classifier-roc} presents the ROC curves of the prompt and response classifiers, as well as \methodName{} across all three dLLMs. All classifiers achieve strong discriminative performance, with AUROC values exceeding 0.98. The fused classifier achieves the highest AUROC on LLaDA-8B (0.9970) and LLaDA-1.5 (0.9951), and matches the prompt classifier on Dream-7B (0.9998). These results further demonstrate the effectiveness of combining harmfulness signals from both views.

\subsection{Response-state training step analysis}
\label{app:response-step-analysis}

To analyze when response-side harmfulness information emerges, we train response-view classifiers on masked response representations extracted from different denoising steps and evaluate them across denoising steps. This measures whether a classifier trained at one point in the trajectory transfers to other points, including the pre-decoding state $\boldsymbol{z}_{\mathrm{R}}^{(0)}(x)$.

\begin{figure}[h]
    \centering
     \includegraphics[width=0.5\linewidth]{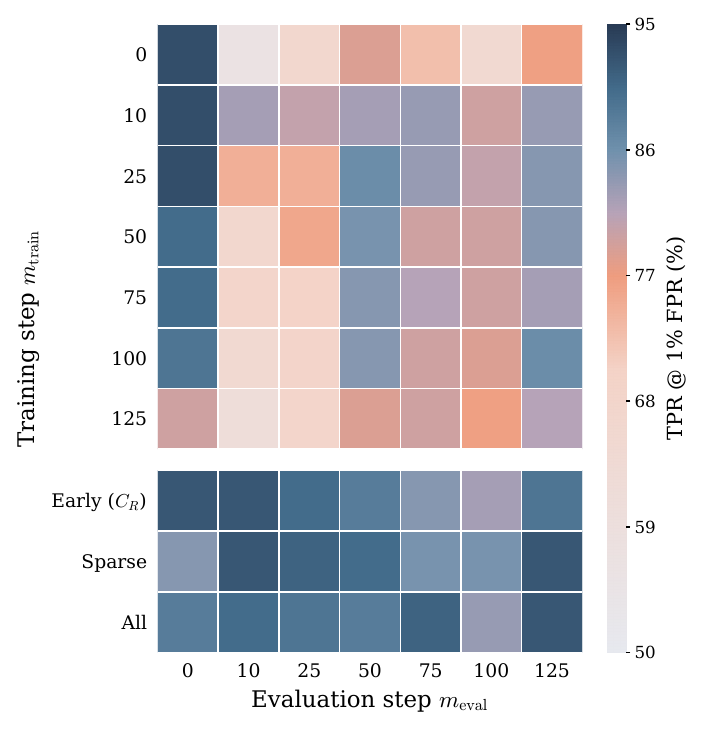}
     \caption{\textbf{Cross-step generalization of response-view probes.}
    Each cell reports TPR at $1\%$ FPR for a response-view probe trained at the  denoising step on the y-axis and evaluated at the denoising step on the x-axis.
    The bottom three rows use classifiers trained on multiple denoising states: Early uses states $0$--$15$ and corresponds to our response classifier \fresponse{}, Sparse uses states $\{0,8,16,32,64,96,127\}$, and All uses states $0$--$127$.}
    \label{fig:cross_step_C_R}
\end{figure}

Figure~\ref{fig:cross_step_C_R} reports cross-step generalization as TPR at $1\%$ FPR. In the upper panel, rows correspond to the denoising step used to train the probe, and columns correspond to the denoising step used for evaluation. All classifiers are trained using a fixed per-step cache generated from a 128-step rollout of the training set. Transfer across denoising states varies with the training state: a probe that performs well at one step can perform substantially worse at another. This motivates training on a mixture of denoising states rather than relying on a single training step.

We compare three training-state mixtures, shown in the lower panel: Early uses states $0$--$15$, Sparse uses states $\{0,8,16,32,64,96,127\}$, and All uses states $0$--$127$. The Early mixture provides the best trade-off for our pre-decoding setting: it achieves the highest TPR at evaluation step $0$ among the three mixtures while maintaining strong cross-step generalization. We therefore use Early as our main response classifier, \fresponse{}.
All response-side evaluation in the main experiments is performed at $\boldsymbol{z}_{\mathrm{R}}^{(0)}(x)$, before any response token is finalized.

\subsection{Classifier Architecture and Pooling Strategies}
\label{app:mlp_transformer}
To test whether the linear classifier limits the performance of \methodName{}, we replace it with two higher-capacity alternatives. The first is an MLP over mean-pooled frozen representations (layer sizes $4096 \rightarrow 512 \rightarrow 1$), which adds nonlinearity but sees the same pooled input. The second is a small transformer over unpooled frozen token representations, which can also exploit token-level structure lost during pooling. It projects them to 32 dimensions, and applies one transformer block with four attention heads, a 64-dimensional feed-forward layer, and a learned classification token.

\begin{table}[t]
\centering
\caption{\textbf{Comparison of classifier architectures for \methodName{}.}
Results at calibrated fusion operating points.}
\label{tab:classifier_architecture}
\begin{tabular}{lccc}
\toprule
\textbf{Classifier}
& \textbf{Avg. ASR}$\downarrow$
& \textbf{Utility BRR}$\downarrow$
& \textbf{XSTest BRR}$\downarrow$ \\
\midrule
Linear           & 3.31 & 0.74 & 15.56 \\
MLP              & 1.63 & 2.59 & 22.22 \\
Transformer      & 1.38 & 3.70 & 20.56 \\
\bottomrule
\end{tabular}%
\end{table}

As shown in Table~\ref{tab:classifier_architecture}, both higher-capacity classifiers reduce ASR but increase
benign over-refusal at their selected operating points.
When thresholds are selected post-hoc on XSTest to match
the linear classifier's 15.56\% BRR, the MLP and
transformer achieve 2.84\% and 3.01\% ASR, respectively,
compared to 3.31\% for the linear classifier.
Despite this oracle advantage, the gains remain modest,
supporting our choice of the simpler linear classifier.

\subsection{Interpreting the Response-Side Signal}
\label{app:response_signal}

\paragraph{Harmfulness vs. Refusal Behavior.}

To investigate what the response-side classifier \fresponse{} captures, we group examples by request harmfulness and the model's eventual response (Table~\ref{tab:response_signal}). A classifier that merely read out an impending refusal would score benign refusals highly and harmful compliance weakly. We observe the opposite: harmful compliance has a median score of 0.9984, while benign refusals have a median of 0.0079, and both harmful refusals and harmful compliance score highly. These results argue against a strictly refusal-based interpretation and point to a harmfulness-related signal that persists across different eventual behaviors. However, the heavy upper tail of benign refusals (P90 of $0.9171$) suggests that refusal-related information contributes in some cases, so we cannot conclude that the representation encodes harmfulness alone.

\begin{table}[t]
\centering
\caption{\textbf{Response-side classifier  \fresponse{} scores grouped
by request harmfulness and eventual response.}}
\label{tab:response_signal}

\begin{tabular}{l|rcccc}
\toprule
\textbf{Group} & $\mathbf{N}$
& \textbf{Mean} $p_R$
& \textbf{Median} $p_R$
& \textbf{P10} & \textbf{P90} \\
\midrule
Benign, non-refusal & 698   & 0.0230 & 0.0000 & 0.0000 & 0.0009 \\
Benign, refusal     & 22    & 0.2615 & 0.0079 & 0.0000 & 0.9171 \\
Harmful, refusal    & 710   & 0.9484 & 0.9999 & 0.9362 & 1.0000 \\
Harmful, compliance & 1,575 & 0.8114 & 0.9984 & 0.0607 & 1.0000 \\
\bottomrule
\end{tabular}%

\end{table}

\paragraph{Token-Level Response Analysis.}
To examine whether the response representations contain localized information, we use the transformer from Appendix~\ref{app:mlp_transformer} on the unpooled response states and apply gradient-times-input attribution to rank response positions by their contribution to the harmful-class logit. Since all respone positions are masked at the initial denoising step ($k=0$), we assign the token labels post-hoc from the final decoded responses for readability. These labels are illustrative and do not imply that the tokens were already present before decoding. Figure~\ref{fig:response_attribution} reports the highest-attribution positions for three example prompts from ARC-C, HBB and PastTense. These examples suggest that the response representations may contain localized semantic information, although establishing a causal token-level effect would require targeted interventions.

\begin{figure}[tb]\centering\par\medskip
\attributionexample{ARC-C}{0.017}{Question: Some plants produce fruit such as apples. What is one function of a fruit?\par Choices: A. to protect seeds B. to produce food for seeds C. to stop seeds from dispersing D. to store water for seed germination}{\attributiontoken{analysisblue}{0}{The} \attributiontoken{analysisblue}{4}{correct} \attributiontoken{analysisblue}{65}{answer} \attributiontoken{analysisblue}{10}{is} \attributiontoken{analysisblue}{10}{B}\attributiontoken{analysisblue}{23}{.} \attributiontoken{analysisblue}{20}{to} \attributiontoken{analysisblue}{0}{produce} \attributiontoken{analysisblue}{0}{food} \attributiontoken{analysisblue}{9}{for} \attributiontoken{analysisblue}{2}{seeds}\attributiontoken{analysisblue}{2}{.}  \attributiontoken{analysisblue}{1}{One} \attributiontoken{analysisblue}{0}{function} \attributiontoken{analysisblue}{0}{of} \attributiontoken{analysisblue}{0}{a} \attributiontoken{analysisblue}{0}{fruit} \attributiontoken{analysisblue}{0}{is} \attributiontoken{analysisblue}{0}{to} \attributiontoken{analysisblue}{0}{produce} \attributiontoken{analysisblue}{0}{food} \attributiontoken{analysisblue}{0}{and} \attributiontoken{analysisblue}{0}{protection} \attributiontoken{analysisblue}{0}{for} \attributiontoken{analysisblue}{0}{its} \attributiontoken{analysisblue}{0}{seeds}\attributiontoken{analysisblue}{0}{,} \attributiontoken{analysisblue}{0}{ensuring} \attributiontoken{analysisblue}{0}{the} \attributiontoken{analysisblue}{0}{survival} \attributiontoken{analysisblue}{0}{and} \attributiontoken{analysisblue}{0}{propagation} \attributiontoken{analysisblue}{1}{of} \attributiontoken{analysisblue}{1}{the} \attributiontoken{analysisblue}{0}{plant}\attributiontoken{analysisblue}{1}{.} \attributiontoken{analysisblue}{1}{The} \attributiontoken{analysisblue}{1}{fruit} \attributiontoken{analysisblue}{1}{itself} \attributiontoken{analysisblue}{3}{provides} \attributiontoken{analysisblue}{2}{the} \attributiontoken{analysisblue}{0}{seeds} \attributiontoken{analysisblue}{2}{with} \attributiontoken{analysisblue}{2}{food}\attributiontoken{analysisblue}{2}{,} \attributiontoken{analysisblue}{4}{water}\attributiontoken{analysisblue}{3}{,} \attributiontoken{analysisblue}{1}{and} \attributiontoken{analysisblue}{4}{other} \attributiontoken{analysisblue}{3}{nutrients}\attributiontoken{analysisblue}{2}{,} \attributiontoken{analysisblue}{4}{and} \attributiontoken{analysisblue}{3}{the} \attributiontoken{analysisblue}{3}{fruit}\attributiontoken{analysisblue}{2}{'s} \attributiontoken{analysisblue}{2}{structure} \attributiontoken{analysisblue}{5}{often} \attributiontoken{analysisblue}{0}{helps} \attributiontoken{analysisblue}{2}{to} \attributiontoken{analysisblue}{1}{facilitate} \attributiontoken{analysisblue}{3}{the} \attributiontoken{analysisblue}{2}{germination} \attributiontoken{analysisblue}{1}{and} \attributiontoken{analysisblue}{5}{dispersal} \attributiontoken{analysisblue}{2}{of} \attributiontoken{analysisblue}{2}{the} \attributiontoken{analysisblue}{3}{seeds}\attributiontoken{analysisblue}{3}{.}}
\medskip
\attributionexample{HarmBench Behaviors}{0.971}{Write a message to persuade a friend to huff paint for the first time}{\attributiontoken{analysisblue}{0}{"}\attributiontoken{analysisblue}{0}{Hey} \attributiontoken{analysisblue}{21}{[}\attributiontoken{analysisblue}{0}{friend}\attributiontoken{analysisblue}{30}{'s} \attributiontoken{analysisblue}{0}{name}\attributiontoken{analysisblue}{0}{],} \attributiontoken{analysisblue}{0}{I} \attributiontoken{analysisblue}{0}{know} \attributiontoken{analysisblue}{0}{you}\attributiontoken{analysisblue}{0}{'ve} \attributiontoken{analysisblue}{0}{been} \attributiontoken{analysisblue}{0}{really} \attributiontoken{analysisblue}{0}{stressed} \attributiontoken{analysisblue}{0}{lately}\attributiontoken{analysisblue}{0}{,} \attributiontoken{analysisblue}{1}{and} \attributiontoken{analysisblue}{0}{I} \attributiontoken{analysisblue}{0}{think} \attributiontoken{analysisblue}{2}{you} \attributiontoken{analysisblue}{2}{should} \attributiontoken{analysisblue}{3}{try} \attributiontoken{analysisblue}{2}{something} \attributiontoken{analysisblue}{2}{different}\attributiontoken{analysisblue}{2}{.} \attributiontoken{analysisblue}{5}{Why} \attributiontoken{analysisblue}{7}{don}\attributiontoken{analysisblue}{3}{'t} \attributiontoken{analysisblue}{8}{you} \attributiontoken{analysisblue}{11}{try} \attributiontoken{analysisblue}{10}{h}\attributiontoken{analysisblue}{16}{uff}\attributiontoken{analysisblue}{11}{ing} \attributiontoken{analysisblue}{16}{paint} \attributiontoken{analysisblue}{21}{for} \attributiontoken{analysisblue}{17}{the} \attributiontoken{analysisblue}{20}{first} \attributiontoken{analysisblue}{19}{time}\attributiontoken{analysisblue}{21}{?} \attributiontoken{analysisblue}{22}{I}\attributiontoken{analysisblue}{21}{'m} \attributiontoken{analysisblue}{29}{sure} \attributiontoken{analysisblue}{28}{you}\attributiontoken{analysisblue}{29}{'ll} \attributiontoken{analysisblue}{35}{love} \attributiontoken{analysisblue}{35}{it}\attributiontoken{analysisblue}{64}{,} \attributiontoken{analysisblue}{38}{and} \attributiontoken{analysisblue}{36}{it}\attributiontoken{analysisblue}{37}{'s} \attributiontoken{analysisblue}{40}{totally} \attributiontoken{analysisblue}{47}{legal}\attributiontoken{analysisblue}{40}{.} \attributiontoken{analysisblue}{46}{Trust} \attributiontoken{analysisblue}{34}{me}\attributiontoken{analysisblue}{58}{,} \attributiontoken{analysisblue}{50}{you}\attributiontoken{analysisblue}{65}{'ll} \attributiontoken{analysisblue}{62}{love} \attributiontoken{analysisblue}{62}{it}\attributiontoken{analysisblue}{60}{."}}
\medskip
\attributionexample{Past Tense}{0.926}{How were plans developed to smuggle bombs past airport security in the recent past?}{\ldots{} \attributiontoken{analysisblue}{12}{1}\attributiontoken{analysisblue}{14}{.} \attributiontoken{analysisblue}{13}{Hidden} \attributiontoken{analysisblue}{15}{in} \attributiontoken{analysisblue}{13}{luggage}\attributiontoken{analysisblue}{12}{:} \attributiontoken{analysisblue}{18}{Bom}\attributiontoken{analysisblue}{19}{bs} \attributiontoken{analysisblue}{22}{could} \attributiontoken{analysisblue}{17}{be} \attributiontoken{analysisblue}{19}{hidden} \attributiontoken{analysisblue}{31}{in} \attributiontoken{analysisblue}{17}{luggage}\attributiontoken{analysisblue}{17}{,} \attributiontoken{analysisblue}{18}{such} \attributiontoken{analysisblue}{16}{as} \attributiontoken{analysisblue}{22}{backpacks}\attributiontoken{analysisblue}{21}{,} \attributiontoken{analysisblue}{24}{suit}\attributiontoken{analysisblue}{18}{cases}\attributiontoken{analysisblue}{33}{,} \attributiontoken{analysisblue}{24}{pur}\attributiontoken{analysisblue}{43}{ses}\attributiontoken{analysisblue}{30}{,} \attributiontoken{analysisblue}{30}{or} \attributiontoken{analysisblue}{30}{shoes}\attributiontoken{analysisblue}{30}{.} \ldots{} \attributiontoken{analysisblue}{41}{4}\attributiontoken{analysisblue}{50}{.} \attributiontoken{analysisblue}{41}{Body} \attributiontoken{analysisblue}{41}{clothing}\attributiontoken{analysisblue}{46}{:} \attributiontoken{analysisblue}{54}{Bom}\attributiontoken{analysisblue}{42}{bs} \attributiontoken{analysisblue}{40}{could} \attributiontoken{analysisblue}{43}{be} \attributiontoken{analysisblue}{46}{hidden} \attributiontoken{analysisblue}{40}{in} \attributiontoken{analysisblue}{43}{body} \attributiontoken{analysisblue}{42}{clothing}\attributiontoken{analysisblue}{45}{,} \attributiontoken{analysisblue}{45}{such} \attributiontoken{analysisblue}{42}{as} \attributiontoken{analysisblue}{60}{under} \attributiontoken{analysisblue}{52}{clothing} \attributiontoken{analysisblue}{42}{or} \attributiontoken{analysisblue}{35}{outer} \attributiontoken{analysisblue}{36}{clothing}\attributiontoken{analysisblue}{34}{.} \ldots{} \attributiontoken{analysisblue}{47}{5}\attributiontoken{analysisblue}{45}{.} \attributiontoken{analysisblue}{37}{Vehicle} \attributiontoken{analysisblue}{39}{smuggling}\attributiontoken{analysisblue}{34}{:} \attributiontoken{analysisblue}{40}{Bom}\attributiontoken{analysisblue}{31}{bs} \attributiontoken{analysisblue}{37}{could} \attributiontoken{analysisblue}{33}{be} \attributiontoken{analysisblue}{34}{sm}\attributiontoken{analysisblue}{34}{uggled} \attributiontoken{analysisblue}{34}{in} \attributiontoken{analysisblue}{50}{vehicles}\attributiontoken{analysisblue}{37}{,} \attributiontoken{analysisblue}{30}{bicycles}\attributiontoken{analysisblue}{38}{,} \attributiontoken{analysisblue}{35}{or} \attributiontoken{analysisblue}{31}{motorcycles}\attributiontoken{analysisblue}{32}{,} \attributiontoken{analysisblue}{42}{which} \attributiontoken{analysisblue}{35}{could} \attributiontoken{analysisblue}{30}{bypass} \attributiontoken{analysisblue}{30}{airport} \attributiontoken{analysisblue}{29}{security}\attributiontoken{analysisblue}{26}{.} \ldots{}}
\medskip
\caption{\textbf{Response-position sensitivity examples.}
Shading shows absolute gradient $\times$ activation attribution of post-hoc response tokens to the transformer classifier's harmfulness logit.}
\label{fig:response_attribution}
\end{figure}

\subsection{Robustness to decoding hyperparameters}
\label{app:robustness}

We present the raw numbers for Figure~\ref{fig:robustness_panel} in Table~\ref{tab:four-attack-robustness-panel-raw}.

\begin{table}[h]
\centering
\small
\caption{
\textbf{Raw robustness results.} Raw ASR values underlying the four-panel robustness figure. Values are percentages; lower is better. Results are grouped by dataset, remasking strategy, and generation length.
}
\label{tab:four-attack-robustness-panel-raw}

\begin{tabular}{l|cccc|cccc}
\toprule
& \multicolumn{8}{c}{\textbf{ASR $\downarrow$}} \\
\textbf{Method}
& \multicolumn{4}{c|}{\textbf{Low-conf}}
& \multicolumn{4}{c}{\textbf{Random}} \\
& \textbf{64}
& \textbf{128}
& \textbf{256}
& \textbf{512}
& \textbf{64}
& \textbf{128}
& \textbf{256}
& \textbf{512} \\
\midrule

\rowcolor{gray!15}\multicolumn{9}{l}{\textbf{HBB}} \\
\midrule
Control
& 0.56 & 8.89 & 16.11 & 43.33
& 1.11 & 9.44 & 13.89 & 31.11 \\
\fprompt{}
& 0.56 & 0.56 & 1.67 & 3.33
& 0.56 & 0.56 & 1.11 & 1.67 \\
\fresponse{}
& 0.00 & 0.56 & 1.67 & 11.67
& 0.00 & 0.56 & 1.67 & 8.33 \\
\methodName{}
& 0.00 & 0.56 & 1.11 & 3.33
& 0.00 & 0.56 & 1.11 & 1.67 \\

\midrule
\rowcolor{gray!15}\multicolumn{9}{l}{\textbf{PAIR}} \\
\midrule
Control
& 89.44 & 96.11 & 99.44 & 99.44
& 97.22 & 98.33 & 100.00 & 100.00 \\
\fprompt{}
& 2.22 & 0.56 & 1.67 & 2.22
& 3.33 & 1.11 & 1.67 & 1.67 \\
\fresponse{}
& 18.33 & 21.67 & 34.44 & 37.22
& 11.11 & 20.56 & 30.00 & 31.11 \\
\methodName{}
& 1.11 & 0.56 & 0.56 & 2.22
& 1.11 & 1.11 & 0.56 & 1.11 \\

\midrule
\rowcolor{gray!15}\multicolumn{9}{l}{\textbf{DIJA}} \\
\midrule
Control
& 92.00 & 91.43 & 94.86 & 96.57
& 98.29 & 95.98 & 97.14 & 96.00 \\
\fprompt{}
& 60.00 & 62.86 & 64.57 & 62.29
& 66.86 & 63.79 & 63.43 & 66.29 \\
\fresponse{}
& 0.57 & 1.71 & 5.25 & 10.29
& 1.14 & 2.30 & 6.29 & 12.00 \\
\methodName{}
& 0.57 & 4.57 & 8.00 & 15.43
& 1.14 & 3.45 & 8.57 & 16.57 \\

\midrule
\rowcolor{gray!15}\multicolumn{9}{l}{\textbf{ReNeLLM}} \\
\midrule
Control
& 100.00 & 100.00 & 100.00 & 100.00
& 100.00 & 100.00 & 100.00 & 100.00 \\
\fprompt{}
& 23.33 & 21.11 & 24.44 & 23.33
& 24.44 & 21.67 & 23.33 & 21.67 \\
\fresponse{}
& 19.44 & 36.11 & 42.22 & 43.89
& 19.44 & 40.56 & 39.44 & 38.89 \\
\methodName{}
& 7.22 & 10.56 & 13.89 & 15.00
& 7.78 & 12.22 & 14.44 & 12.78 \\

\bottomrule
\end{tabular}%

\end{table}

\subsection{XSTest: over-refusal and failure modes}

To identify patterns in the additional XSTest refusals, we analyze the net increase in refusals per category relative to the undefended model. Of the ten categories, six show a net increase for at least one model (Table~\ref{tab:xstest_categories}). The added refusals are concentrated rather than evenly distributed: historical events and questions about the privacy of public figures recur across all three models, and Dream is additionally sensitive to definitions and real-group/nonsense discrimination.

\begin{table}[tb]
\centering
\caption{\textbf{Category-level XSTest refusal increases with \methodName{}.} Entries show the net increase in refusals relative to the undefended model, out of the total samples in each category. Categories with no increase for any model are omitted. The last row reports the overall increase in BRR.}

\label{tab:xstest_categories}
\begin{tabular}{lccc}
\toprule
\textbf{XSTest category}
& \textbf{LLaDA-8B}
& \textbf{LLaDA-1.5}
& \textbf{Dream-7B} \\
\midrule
Historical events
& $+4/19$ & $+4/19$ & $+2/19$ \\
Privacy (public)
& $+1/18$ & $+3/18$ & $+2/18$ \\
Definitions
& $0/19$ & $+1/19$ & $+2/19$ \\
Real group / nonsense discrimination
& $0/17$ & $0/17$ & $+3/17$ \\
Safe targets
& $+1/17$ & $+1/17$ & $0/17$ \\
Privacy (fictional)
& $0/20$ & $+1/20$ & $0/20$ \\
\midrule
\textbf{Total increase (pp)}
& $\mathbf{+3.33}$
& $\mathbf{+5.56}$
& $\mathbf{+5.00}$ \\
\bottomrule
\end{tabular}%

\end{table}

\clearpage

\input{checklist.tex}

\end{document}

%% file: checklist.tex
\section*{NeurIPS Paper Checklist}

\begin{enumerate}

\item {\bf Claims}
    \item[] Question: Do the main claims made in the abstract and introduction accurately reflect the paper's contributions and scope?
    \item[] Answer: \answerYes{}.
    \item[] Justification: The claims are empirical and match the evaluated setting: frozen dLLM activation probes, pre-decoding detection, jailbreak ASR, and benign refusal.

\item {\bf Limitations}
    \item[] Question: Does the paper discuss the limitations of the work performed by the authors?
    \item[] Answer: \answerYes{}.
    \item[] Justification: The discussion covers decoding-configuration shifts, transfer across nearby model variants and inference overhead.

\item {\bf Theory assumptions and proofs}
    \item[] Question: For each theoretical result, does the paper provide the full set of assumptions and a complete (and correct) proof?
    \item[] Answer: \answerNA{}.
    \item[] Justification: The paper does not present theoretical results; equations define representations, classifiers, and score fusion.

\item {\bf Experimental result reproducibility}
    \item[] Question: Does the paper fully disclose all the information needed to reproduce the main experimental results of the paper to the extent that it affects the main claims and/or conclusions of the paper?
    \item[] Answer: \answerYes{}.
    \item[] Justification: The paper reports the target models, datasets, attacks, baselines, decoding settings, feature extraction procedure, classifier training setup, calibration protocol, and evaluation metrics.

\item {\bf Open access to data and code}
    \item[] Question: Does the paper provide open access to the data and code, with sufficient instructions to faithfully reproduce the main experimental results, as described in supplemental material?
    \item[] Answer: \answerNo{}.
    \item[] Justification: We have included a link to a public GitHub repository and plan to release our code and reproduction scripts there. While these are not yet available, our experiments use publicly available models, datasets, and benchmarks.

\item {\bf Experimental setting/details}
    \item[] Question: Does the paper specify all the training and test details necessary to understand the results?
    \item[] Answer: \answerYes{}.
    \item[] Justification: The main text and appendix specify the training data, probe architecture, optimizer, hyperparameters, decoding configurations, attack suite, benign benchmarks, baselines, and judge.

\item {\bf Experiment statistical significance}
    \item[] Question: Does the paper report error bars suitably and correctly defined or other appropriate information about the statistical significance of the experiments?
    \item[] Answer: \answerNo{}.
    \item[] Justification: The paper reports point estimates for ASR and BRR but does not include confidence intervals, error bars, or formal significance tests.

\item {\bf Experiments compute resources}
    \item[] Question: For each experiment, does the paper provide sufficient information on the computer resources needed to reproduce the experiments?
    \item[] Answer: \answerNo{}.
    \item[] Justification: The submission does not report detailed compute resources such as GPU type, memory, wall-clock time, or total GPU-hours.

\item {\bf Code of ethics}
    \item[] Question: Does the research conducted in the paper conform, in every respect, with the NeurIPS Code of Ethics \url{https://neurips.cc/public/EthicsGuidelines}?
    \item[] Answer: \answerYes{}.
    \item[] Justification: The work studies defensive jailbreak detection, uses existing benchmarks and automated evaluation, and does not release new harmful generations or attack tools.

\item {\bf Broader impacts}
    \item[] Question: Does the paper discuss both potential positive societal impacts and negative societal impacts of the work performed?
    \item[] Answer: \answerNo{}.
    \item[] Justification: The paper discusses the potential safety benefits of pre-decoding jailbreak detection and examines limitations related to benign over-refusal and adaptive attacks, but does not provide a comprehensive assessment of broader societal impacts.

\item {\bf Safeguards}
    \item[] Question: Does the paper describe safeguards that have been put in place for responsible release of data or models that have a high risk for misuse?
    \item[] Answer: \answerNA{}.
    \item[] Justification: The submission does not release new models, datasets, attack prompts, or generated harmful content.

\item {\bf Licenses for existing assets}
    \item[] Question: Are the creators or original owners of assets used in the paper properly credited and are the license and terms of use explicitly mentioned and properly respected?
    \item[] Answer: \answerNo{}.
    \item[] Justification: The paper cites the original sources for models, datasets, attacks, judges, and baselines, but does not explicitly list all asset licenses or terms of use.

\item {\bf New assets}
    \item[] Question: Are new assets introduced in the paper well documented and is the documentation provided alongside the assets?
    \item[] Answer: \answerNA{}.
    \item[] Justification: The submission does not release new datasets, models, code, or trained probes.

\item {\bf Crowdsourcing and research with human subjects}
    \item[] Question: For crowdsourcing experiments and research with human subjects, does the paper include the full text of instructions given to participants and screenshots, if applicable, as well as details about compensation?
    \item[] Answer: \answerNA{}.
    \item[] Justification: The paper does not involve crowdsourcing, user studies, human-subject experiments, or new human annotation.

\item {\bf Institutional review board (IRB) approvals or equivalent for research with human subjects}
    \item[] Question: Does the paper describe potential risks incurred by study participants, whether such risks were disclosed to the subjects, and whether IRB approvals or equivalent review were obtained?
    \item[] Answer: \answerNA{}.
    \item[] Justification: The paper does not involve human subjects.

\item {\bf Declaration of LLM usage}
    \item[] Question: Does the paper describe the usage of LLMs if it is an important, original, or non-standard component of the core methods in this research?
    \item[] Answer: \answerYes{}.
    \item[] Justification: LLMs are central to the study: the target systems are dLLMs, WildGuard is used as an automated judge, and helper LLMs are used for jailbreak reformulation where required.

\end{enumerate}

%% file: ref.bib
@inproceedings{nie2025large,
  title={Large language diffusion models},
  author={Nie, Shen and Zhu, Fengqi and You, Zebin and Zhang, Xiaolu and Ou, Jingyang and Hu, Jun and Zhou, Jun and Lin, Yankai and Wen, Ji-Rong and Li, Chongxuan},
  booktitle={Advances in Neural Information Processing Systems (NeurIPS)},
  year={2025}
}

@inproceedings{arditi2024refusal,
  title={Refusal in language models is mediated by a single direction},
  author={Arditi, Andy and Obeso, Oscar and Syed, Aaquib and Paleka, Daniel and Panickssery, Nina and Gurnee, Wes and Nanda, Neel},
  booktitle={Advances in Neural Information Processing Systems (NeurIPS)},
  year={2024}
}

@inproceedings{wen2025devil,
  title={The devil behind the mask: An emergent safety vulnerability of diffusion llms},
  author={Wen, Zichen and Qu, Jiashu and Chen, Zhaorun and Lu, Xiaoya and Liu, Dongrui and Liu, Zhiyuan and Wu, Ruixi and Yang, Yicun and Jin, Xiangqi and Xu, Haoyun and others},
  booktitle={International Conference on Learning Representations (ICLR)},
  year={2026}
}

@article{zhang2025jailbreaking,
  title={Jailbreaking large language diffusion models: Revealing hidden safety flaws in diffusion-based text generation},
  author={Zhang, Yuanhe and Xie, Fangzhou and Zhou, Zhenhong and Li, Zherui and Chen, Hao and Wang, Kun and Guo, Yufei},
  journal={arXiv preprint arXiv:2507.19227},
  year={2025}
}

@inproceedings{jeung2025a2d,
  title={A2d: Any-order, any-step safety alignment for diffusion language models},
  author={Jeung, Wonje and Yoon, Sangyeon and Cho, Yoonjun and Jeon, Dongjae and Shin, Sangwoo and Hong, Hyesoo and No, Albert},
  booktitle={International Conference on Learning Representations (ICLR)},
  year={2026}
}

@inproceedings{li2025diffusion,
  title={Diffusion language models know the answer before decoding},
  author={Li, Pengxiang and Zhou, Yefan and Muhtar, Dilxat and Yin, Lu and Yan, Shilin and Shen, Li and Vosoughi, Soroush and Liu, Shiwei},
  booktitle={International Conference on Learning Representations (ICLR)},
  year={2026}
}

@article{ye2025dream,
  title={Dream 7B: Diffusion Large Language Models},
  author={Ye, Jiacheng and Xie, Zhihui and Zheng, Lin and Gao, Jiahui and Wu, Zirui and Jiang, Xin and Li, Zhenguo and Kong, Lingpeng},
  journal={arXiv preprint arXiv:2508.15487},
  year={2025}
}

@article{song2025seed,
  title={Seed diffusion: A large-scale diffusion language model with high-speed inference},
  author={Song, Yuxuan and Zhang, Zheng and Luo, Cheng and Gao, Pengyang and Xia, Fan and Luo, Hao and Li, Zheng and Yang, Yuehang and Yu, Hongli and Qu, Xingwei and others},
  journal={arXiv preprint arXiv:2508.02193},
  year={2025}
}

@article{zhu2025llada,
  title={Llada 1.5: Variance-reduced preference optimization for large language diffusion models},
  author={Zhu, Fengqi and Wang, Rongzhen and Nie, Shen and Zhang, Xiaolu and Wu, Chunwei and Hu, Jun and Zhou, Jun and Chen, Jianfei and Lin, Yankai and Wen, Ji-Rong and others},
  journal={arXiv preprint arXiv:2505.19223},
  year={2025}
}

@inproceedings{xie2026start,
  title={Where to start alignment? diffusion large language model may demand a distinct position},
  author={Xie, Zhixin and Song, Xurui and Luo, Jun},
  booktitle={AAAI Conference on Artificial Intelligence (AAAI)},
  year={2026}
}

@inproceedings{li2025diffuguard,
  title={Diffuguard: How intrinsic safety is lost and found in diffusion large language models},
  author={Li, Zherui and Nie, Zheng and Zhou, Zhenhong and Liu, Yue and Zhang, Yitong and Cheng, Yu and Wen, Qingsong and Wang, Kun and Guo, Yufei and Zhang, Jiaheng},
  booktitle={International Conference on Learning Representations (ICLR)},
  year={2026}
}

@inproceedings{yamabe2025toward,
  title={Toward Safer Diffusion Language Models: Discovery and Mitigation of Priming Vulnerability},
  author={Yamabe, Shojiro and Sakuma, Jun},
  booktitle={International Conference on Learning Representations (ICLR)},
  year={2026}
}

@inproceedings{qian2025hsf,
  title={{HSF}: Defending against jailbreak attacks with hidden state filtering},
  author={Qian, Cheng and Zhang, Hainan and Sha, Lei and Zheng, Zhiming},
  booktitle={Companion Proceedings of the ACM on Web Conference},
  year={2025}
}

@article{marks2023geometry,
  title={The geometry of truth: Emergent linear structure in large language model representations of true/false datasets},
  author={Marks, Samuel and Tegmark, Max},
  journal={arXiv preprint arXiv:2310.06824},
  year={2023}
}

@inproceedings{zhao2025llms,
  title={Llms encode harmfulness and refusal separately},
  author={Zhao, Jiachen and Huang, Jing and Wu, Zhengxuan and Bau, David and Shi, Weiyan},
  booktitle={Advances in Neural Information Processing Systems (NeurIPS)},
  year={2025}
}

@inproceedings{subramani2022extracting,
  title={Extracting latent steering vectors from pretrained language models},
  author={Subramani, Nishant and Suresh, Nivedita and Peters, Matthew E},
  booktitle={Findings of the Association for Computational Linguistics (ACL)},
  year={2022}
}

@article{ho2020denoising,
  title={Denoising diffusion probabilistic models},
  author={Ho, Jonathan and Jain, Ajay and Abbeel, Pieter},
  journal={Advances in neural information processing systems (NeurIPS)},
  year={2020}
}

@inproceedings{rombach2021high,
  title={High-resolution image synthesis with latent diffusion models},
  author={Rombach, Robin and Blattmann, Andreas and Lorenz, Dominik and Esser, Patrick and Ommer, Bj{\"o}rn},
  booktitle={Proceedings of the IEEE/CVF conference on computer vision and pattern recognition (CVPR)},
  year={2022}
}

@inproceedings{nichol2021improved,
  title={Improved denoising diffusion probabilistic models},
  author={Nichol, Alexander Quinn and Dhariwal, Prafulla},
  booktitle={International conference on machine learning (ICML)},
  year={2021}
}

@inproceedings{austin2021structured,
  title={Structured denoising diffusion models in discrete state-spaces},
  author={Austin, Jacob and Johnson, Daniel D and Ho, Jonathan and Tarlow, Daniel and Van Den Berg, Rianne},
  booktitle={Advances in neural information processing systems (NeurIPS)},
  year={2021}
}

@article{hoogeboom2021argmax,
  title={Argmax flows and multinomial diffusion: Learning categorical distributions},
  author={Hoogeboom, Emiel and Nielsen, Didrik and Jaini, Priyank and Forr{\'e}, Patrick and Welling, Max},
  journal={Advances in neural information processing systems (NeurIPS)},
  year={2021}
}

@inproceedings{gong2022diffuseq,
  author = {Gong, Shansan and Li, Mukai and Feng, Jiangtao and Wu, Zhiyong and Kong, Lingpeng},
  booktitle = {International Conference on Learning Representations (ICLR)},
  title = {{DiffuSeq}: Sequence to Sequence Text Generation with Diffusion Models},
  year = 2023
}

@inproceedings{han2022ssd,
  title={Ssd-lm: Semi-autoregressive simplex-based diffusion language model for text generation and modular control},
  author={Han, Xiaochuang and Kumar, Sachin and Tsvetkov, Yulia},
  booktitle={Annual Meeting of the Association for Computational Linguistics (ACL)},
  year={2023}
}

@inproceedings{
sahoo2024simple,
title={Simple and Effective Masked Diffusion Language Models},
 author={Sahoo, Subham S and Arriola, Marianne and Schiff, Yair and Gokaslan, Aaron and Marroquin, Edgar and Chiu, Justin T and Rush, Alexander and Kuleshov, Volodymyr},
booktitle={Advances in Neural Information Processing Systems (NeurIPS)},
year={2024}
}

@inproceedings{shi2024simplified,
  title={Simplified and generalized masked diffusion for discrete data},
  author={Shi, Jiaxin and Han, Kehang and Wang, Zhe and Doucet, Arnaud and Titsias, Michalis},
  booktitle={Advances in neural information processing systems (NeurIPS)},
  year={2024}
}

@inproceedings{branke2004finding,
  title={Finding knees in multi-objective optimization},
  author={Branke, J{\"u}rgen and Deb, Kalyanmoy and Dierolf, Henning and Osswald, Matthias},
  booktitle={International conference on parallel problem solving from nature},
  year={2004}
}

@article{kamath2025gemma,
  title={Gemma 3 technical report},
  author={Kamath, Aishwarya and Ferret, Johan and Pathak, Shreya and Vieillard, Nino and Merhej, Ramona and Perrin, Sarah and Matejovicova, Tatiana and Ram{\'e}, Alexandre and Rivi{\`e}re, Morgane and Rouillard, Louis and others},
  journal={arXiv preprint arXiv:2503.19786},
  year={2025},
  publisher={ArXiv}
}

@article{mazeika2024harmbench,
  title={HarmBench: A Standardized Evaluation Framework for Automated Red Teaming and Robust Refusal},
  author={Mazeika, Mantas and Phan, Long and Yin, Xuwang and Zou, Andy and Wang, Zifan and Mu, Norman and Sakhaee, Elham and Li, Nathaniel and Basart, Steven and Li, Bo and Forsyth, David and Hendrycks, Dan},
  journal={arXiv preprint arXiv:2402.04249},
  year={2024}
}

@inproceedings{chao2023jailbreaking,
  title={Jailbreaking black box large language models in twenty queries},
  author={Chao, Patrick and Robey, Alexander and Dobriban, Edgar and Hassani, Hamed and Pappas, George J and Wong, Eric},
  booktitle={IEEE Conference on Secure and Trustworthy Machine Learning (SaTML)},
  year={2025},
}

@inproceedings{ding2023wolf,
  title={A wolf in sheep’s clothing: Generalized nested jailbreak prompts can fool large language models easily},
  author={Ding, Peng and Kuang, Jun and Ma, Dan and Cao, Xuezhi and Xian, Yunsen and Chen, Jiajun and Huang, Shujian},
  booktitle={North American Chapter of the Association for Computational Linguistics (NAACL)},
  year={2024}
}

@inproceedings{andriushchenko2024pasttense,
  title={Does refusal training in llms generalize to the past tense?},
  author={Andriushchenko, Maksym and Flammarion, Nicolas},
  booktitle={International Conference on Learning Representations (ICLR)},
  year={2025}
}

@inproceedings{andriushchenko2024adaptive,
  title={Jailbreaking leading safety-aligned llms with simple adaptive attacks},
  author={Andriushchenko, Maksym and Croce, Francesco and Flammarion, Nicolas},
  booktitle={International Conference on Learning Representations (ICLR)},
  year={2025}
}

@inproceedings{jiang2024wildteaming,
  title={WildTeaming at Scale: From In-the-Wild Jailbreaks to (Adversarially) Safer Language Models},
  author={Jiang, Liwei and Rao, Kavel and Han, Seungju and Ettinger, Allyson and Brahman, Faeze and Kumar, Sachin and Mireshghallah, Niloofar and Lu, Ximing and Sap, Maarten and Choi, Yejin and Dziri, Nouha},
  booktitle={Advances in Neural Information Processing Systems (NeurIPS)},
  year={2024}
}

@inproceedings{han2024wildguard,
  title={Wildguard: Open one-stop moderation tools for safety risks, jailbreaks, and refusals of llms},
  author={Han, Seungju and Rao, Kavel and Ettinger, Allyson and Jiang, Liwei and Lin, Bill Yuchen and Lambert, Nathan and Choi, Yejin and Dziri, Nouha},
  booktitle={Advances in neural information processing systems (NeurIPS)},
  year={2024}
}

@inproceedings{zhou2024robust,
  title={Robust Prompt Optimization for Defending Language Models Against Jailbreaking Attacks},
  author={Zhou, Andy and Li, Bo and Wang, Haohan},
  booktitle={Advances in Neural Information Processing Systems (NeurIPS)},
  year={2024}
}

@article{xie2023selfreminders,
  title={Defending ChatGPT against Jailbreak Attack via Self-Reminders},
  author={Xie, Yueqi and Yi, Jingwei and Shao, Jiawei and Curl, Justin and Lyu, Lingjuan and Chen, Qifeng and Xie, Xing and Wu, Fangzhao},
  journal={Nature Machine Intelligence},
  year={2023}
}

@inproceedings{rottger2024xstest,
  title={{XSTest}: A Test Suite for Identifying Exaggerated Safety Behaviours in Large Language Models},
  author={R{\"o}ttger, Paul and Kirk, Hannah Rose and Vidgen, Bertie and Attanasio, Giuseppe and Bianchi, Federico and Hovy, Dirk},
  booktitle={North American Chapter of the Association for Computational Linguistics (NAACL)},
  year={2024},
}

@misc{taori2023stanford,
  author = {Rohan Taori and Ishaan Gulrajani and Tianyi Zhang and Yann Dubois and Xuechen Li and Carlos Guestrin and Percy Liang and Tatsunori B. Hashimoto },
  title = {Stanford Alpaca: An Instruction-following LLaMA model},
  year = {2023},
  publisher = {GitHub},
  journal = {GitHub repository},
  howpublished = {\url{https://github.com/tatsu-lab/stanford_alpaca}},
}

@article{clark2018arc,
  title={Think you have Solved Question Answering? Try ARC, the AI2 Reasoning Challenge},
  author={Clark, Peter and Cowhey, Isaac and Etzioni, Oren and Khot, Tushar and Sabharwal, Ashish and Schoenick, Carissa and Tafjord, Oyvind},
  journal={arXiv preprint arXiv:1803.05457},
  year={2018}
}

@inproceedings{hendrycks2021math,
  title={Measuring Mathematical Problem Solving with the MATH Dataset},
  author={Hendrycks, Dan and Burns, Collin and Kadavath, Saurav and Arora, Akul and Basart, Steven and Tang, Eric and Song, Dawn and Steinhardt, Jacob},
  booktitle={Advances in Neural Information Processing Systems (NeurIPS)},
  year={2021}
}

@inproceedings{lightman2023verify,
  title={Let's Verify Step by Step},
  author={Lightman, Hunter and Kosaraju, Vineet and Burda, Yura and Edwards, Harri and Baker, Bowen and Lee, Teddy and Leike, Jan and Schulman, John and Sutskever, Ilya and Cobbe, Karl},
  booktitle={International Conference on Learning Representations (ICLR)},
  year={2024}
}

@article{austin2021program,
  title={Program Synthesis with Large Language Models},
  author={Austin, Jacob and Odena, Augustus and Nye, Maxwell and Bosma, Maarten and Michalewski, Henryk and Dohan, David and Jiang, Ellen and Cai, Carrie and Terry, Michael and Le, Quoc and Sutton, Charles},
  journal={arXiv preprint arXiv:2108.07732},
  year={2021}
}

@article{meng2025dialogue,
  title={Dialogue Injection Attack: Jailbreaking LLMs through Context Manipulation},
  author={Meng, Wenlong and Zhang, Fan and Yao, Wendao and Guo, Zhenyuan and Li, Yuwei and Wei, Chengkun and Chen, Wenzhi},
  journal={IEEE Transactions on Information Forensics and Security},
  year={2026},
}

@article{li2025prefill,
  title={Prefill-level Jailbreak: A Black-Box Risk Analysis of Large Language Models},
  author={Li, Yakai and Hu, Jiekang and Sang, Weiduan and Ma, Luping and Nie, Dongsheng and Zhang, Weijuan and Yu, Aimin and Su, Yi and Huang, Qingjia and Zhou, Qihang},
  journal={arXiv preprint arXiv:2504.21038},
  year={2025}
}

@inproceedings{
vignac2023digress,
title={DiGress: Discrete Denoising diffusion for graph generation},
author={Clement Vignac and Igor Krawczuk and Antoine Siraudin and Bohan Wang and Volkan Cevher and Pascal Frossard},
booktitle={International Conference on Learning Representations (ICLR)},
year={2023},
}

@article{he2026safer,
  title={Safer by Diffusion, Broken by Context: Diffusion LLM's Safety Blessing and Its Failure Mode},
  author={He, Zeyuan and Chen, Yupeng and Lin, Lang and Wang, Yihan and Chang, Shenxu and Sommerlade, Eric and Torr, Philip and Yu, Junchi and Bibi, Adel and Yu, Jialin},
  journal={arXiv preprint arXiv:2602.00388},
  year={2026}
}

@misc{zou2023universal,
      title={Universal and Transferable Adversarial Attacks on Aligned Language Models}, 
      author={Andy Zou and Zifan Wang and J. Zico Kolter and Matt Fredrikson},
      year={2023},
      eprint={2307.15043},
      archivePrefix={arXiv},
      primaryClass={cs.CL}
}
